\documentclass[runningheads]{llncs}

\usepackage[T1]{fontenc}
\usepackage{graphicx}
\usepackage{adjustbox}
\usepackage{multirow}
\usepackage{amsmath}
\usepackage{tabularx}
\usepackage{booktabs}
\newcolumntype{C}{>{\centering\arraybackslash}X}
\newcolumntype{R}{>{\raggedleft\arraybackslash}X}
\newcolumntype{L}{>{\raggedright\arraybackslash}X}

\usepackage{algorithm}
\usepackage{algpseudocode}
\usepackage{comment}
\usepackage{float}
\newfloat{algorithm}{t}{lop}
\algnewcommand\algorithmicsymbols{\textbf{Symbols:}}
\algnewcommand\ASymbols{\item[\algorithmicsymbols]}
\algnewcommand\algorithmicinput{\textbf{Input:}}
\algnewcommand\algorithmicoutput{\textbf{Output:}}
\algnewcommand\Input{\item[\algorithmicinput]}
\algnewcommand\Output{\item[\algorithmicoutput]}
\algnewcommand\algorithmicforeach{\textbf{for each}}
\algdef{S}[FOR]{ForEach}[1]{\algorithmicforeach\ #1\ \algorithmicdo}
\algtext*{EndIf}
\usepackage{makecell}
\usepackage{graphicx}
\usepackage{subcaption}
\usepackage{xcolor}

\usepackage{footnote}
\usepackage[hybrid]{markdown}

\usepackage{amsmath}
\usepackage{amsfonts}
\usepackage{amssymb}

\usepackage{bibnames}

\begin{document}

\title{\textbf{...}}
\title{\textbf{
    %Evaluation of Multi-objective Learning Algorithms for Imbalanced Data Problems    
    %Alternatywny tytuł
Evaluation of Multi- and Single-objective Learning Algorithms for Imbalanced Data 
}}
\titlerunning{Evaluation of Multi-objective Learning for Imbalanced Data}

\author{
     Szymon Wojciechowski\orcidID{0000-0002-8437-5592} \\
     Michał Woźniak\orcidID{0000-0003-0146-4205}
 }

 \authorrunning{S. Wojciechowski and M.Woźniak}

 \institute{Department of Systems and Computer Networks, \\
 Wrocław University of Science and Technology, \\
 Wybrzeże Wyspiańskiego 27, 50-370 Wrocław, Poland
 \email{szymon.wojciechowski@pwr.edu.pl}}

%\author{
%    Authors anonymized for the review
%}

%\institute{
%    Institute anonymized for the review
%}

%\authorrunning{Authors anonymized for the review}

\maketitle
\begin{abstract}

Many machine learning tasks aim to find models that work well not for a single, but for a group of criteria, often opposing ones. One such example is imbalanced data classification, where, on the one hand, we want to achieve the best possible classification quality for data from the minority class without degrading the classification quality of the majority class. One solution is to propose an aggregate learning criterion and reduce the multi-objective learning task to a single-criteria optimization problem. Unfortunately, such an approach is characterized by ambiguity of interpretation since the value of the aggregated criterion does not indicate the value of the component criteria. Hence, there are more and more proposals for algorithms based on multi-objective optimization (MOO), which can simultaneously optimize multiple criteria. However, such an approach results in a set of multiple non-dominated solutions (Pareto front). The selection of a single solution from the Pareto front is a challenge itself, and much attention is paid to the issue of how to select it considering user preferences, as well as how to compare solutions returned by different MOO algorithms among themselves. Thus, a significant gap has been identified in the classifier evaluation methodology, i.e., how to reliably compare methods returning single solutions with algorithms returning solutions in the form of Pareto fronts.

To fill the aforementioned gap, this article proposes a new, reliable way of evaluating algorithms based on multi-objective algorithms with methods that return single solutions while pointing out solutions from a Pareto front tailored to the user's preferences. This work focuses only on algorithm comparison, not their learning. The algorithms selected for this study are illustrative to help understand the proposed approach.

\end{abstract}

\section{Introduction}

%Machine learning (\textsc{ml}) has emerged as a dominant area of research in advancing artificial intelligence. Learning systems are employed in almost every sector of modern technology and industry, and for many individuals, they have become a routine aspect of their lives.

A machine learning system is a system that performs a particular task using information (experience) from the surrounding environment~%. That description refers to Mitchel's definition 
\cite{mitchell2018}, i.e., %, in which a third essential component is the performance assessment criterion. 
it uses learning data to improve a criterion related to the quality of its performance, such as classification quality.
Inherently, a machine learning algorithm is a heuristic that solves an optimization problem, although it is under problem-specific rules and limitations of the data distribution.

% \color{red}

\begin{figure}
    \centering
    \includegraphics[width=0.75\linewidth]{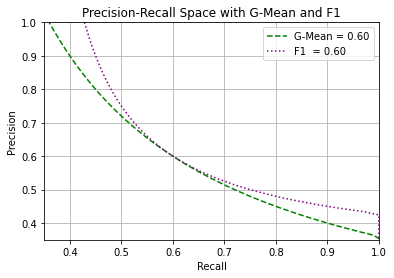}
    \caption{G-mean and $F_1$ curves in the recall-precision space.}
    \label{fig:gmf1}
\end{figure}

We are faced with formulating a correct criterion for evaluating the classification performance for many classifier learning tasks. The probabilistic approach indicates that the best classifier should minimize the so-called overall risk \cite{Duda:2001}. However, practically, we often cannot determine probabilistic characteristics (conditional probability density functions or prior probability) and do not have access to a loss matrix indicating the cost of committing errors between classes. Hence, accuracy is a common metric for the 0-1 loss function. It is sufficient for problems for which the cost of errors for each pair of classes is the same> However, for many tasks, it is not symmetric, causing quality assessment based on accuracy to be biased toward classes with a small error cost. A typical example of such a task is imbalanced data classification, i.e., when the number of objects among classes varies strongly and, as a rule, the cost of errors made on the sparse classes is significantly higher than errors made on the other classes. In this paper, we will consider a binary problem, i.e., we will distinguish minority and majority classes, e.g., fraud detection, where the minority class is fraudulent transactions and the majority class is legitimate ones.
Empirical evidence proves that $accuracy$ is strongly biased to favor the majority class and might produce misleading conclusions. 
Thus, to correctly assess the quality of the resulting classifier, we should simultaneously evaluate its quality on both majority and minority classes, e.g., using at least a pair of metrics, such as specificity and sensitivity (recall), or precision and recall. However, many researchers prefer to assess the quality using a single metric, resulting in a search for metrics that aggregate knowledge on positive and negative class performances \cite{Stapor:2021}. Examples of such metrics are the arithmetic, geometric, or harmonic means of \emph{recall} and \emph{precision} (or \emph{specificity}). 
Unfortunately, many people fail to realize that such metrics are ambiguous because the same metric value can be taken for different $precision$ and $recall$ values, as shown in Fig. \ref{fig:gmf1}. Thus, one might suspect that machine learning methods that use such an aggregate metric as a criterion will be biased toward specific values of simple metrics without providing the information that there are equally good solutions (in terms of a given metric) for other values of \emph{precision} and \emph{recall}.

In addition, some of the metrics are de facto parametric (such as $F_{{beta}}$). When evaluating a classifier, one would need to consider several values of its parameter as they express the kind of trade-off between both components and should be appropriately set. Hand \cite{Hand:2018} criticized the common use of $F_1$ metric, i.e., for the parameter $\beta=1$, as the user should set this parameter to express how much more important the $recall$ is than the $precision$. Unfortunately, for many practical problems, access to end-user preferences in this area is unknown. The interesting study of metrics behaviors for the chosen classification task could also be found in \cite{Brzezinski:2019}.

One solution that can be applied to the above problem is to abandon learning to optimize a single aggregate criterion and treat the imbalanced data classifier learning task as a multi-criteria optimization problem. Such an approach allows us to simultaneously optimize a set of criteria (e.g., precision and recall) and the set of solutions so obtained in the form of a Pareto front. 

Such an approach, however, implies the problem of selecting a solution from a pool of non-dominant solutions. It is a non-trivial task and requires the user’s participation, who is either to identify single solutions that best fit requirements or can prioritize individual criteria.
However, a problem that needs to be addressed is comparing solutions obtained from different algorithms. While comparing algorithms based on a single metric is quite widespread, several experimental protocols have been developed, often supported by statistical analysis tools \cite{Stapor:2021,Japkowicz:2024}. Similarly, it is possible to point to a relatively huge pool of methods comparing the results of algorithms based on multi-criteria optimization \cite{laszczyk2019}. Unfortunately, there are no reliable methods to compare methods that return a single solution with methods that return a pool of solutions resulting from multi-criteria learning.

The paper proposes addressing this methodological gap by comparing a classifier learning algorithm that returns a pool of solutions with a single classifier. The application of the proposed method will be demonstrated in a selected case study. The algorithms chosen for this study are for illustrative purposes only, to help understand the proposed approach, and are not new learning method proposals.

\section{Preliminaries}

This paper addresses the problem of imbalanced data classification and how to compare classifiers obtained from traditional algorithms that return a single solution, as well as those that can return pools of solutions through multi-criteria optimization methods. This section introduces the metrics used in this task and briefly introduces multi-criteria learning.

\subsection{Metrics for imbalanced data classifier evaluation}
%A typical model evaluation protocol aims to identify the model with the highest prediction accuracy, supported by the statistical significance of such a test. Quality is measured using metrics based on a confusion matrix in which the rows and columns denote the actual and predicted classes. Comparisons are counted and accumulate in the appropriate cells (see Figure \ref{fig:consufion}), which are essential model characteristics.

A typical experimental protocol for classifier evaluation aims to identify the model with the highest prediction quality, generally using a selected statistical analysis tool. Quality metrics can be counted on the information in the confusion matrix (see Figure \ref{fig:consufion}).

\begin{figure}[H]
\begin{adjustbox}{width=0.65\columnwidth,center}
\begin{tabular}{|cc|c|c|}
  \cline{3-4}
 \multicolumn{2}{c|}{\textsc{ }} & \multicolumn{2}{c|}{\textsc{predicted class}} \\
 \multicolumn{2}{c|}{\textsc{ }} & \textsc{positive} & \textsc{negative}\\
  \hline
    \multirow{6}{*}{\rotatebox[origin=c]{90}{\textsc{true class}}}  &&&\\
    & \textsc{positive} & \emph{True Positive ($TP$)} & \emph{False Negative ($FN$)}\\
   
    &&&\\
    \cline{2-4}
     &&&\\
   & \textsc{negative} & \emph{False Positive ($FP$)} & \emph{True Negative ($TN$)}\\
   &&&\\
    
  \hline
\end{tabular} 
\end{adjustbox}
\caption{Confusion matrix for a two-class problem.}
\label{fig:consufion}
\end{figure}

%The most popular way to assess the model quality is to estimate the \textit{accuracy}, denoted by the percentage of correctly classified objects. However, this approach is sensitive to class imbalance and can project results biased toward the majority class. To overcome such issues, more class-oriented metrics, such as \textit{sensitivity} (TPR), \textit{specificity} (TNR), and \textit{precision} (PPV), can be formulated as in Equation \ref{eq:base}.
The most popular metric is \textit{accuracy}. However, using it when dealing with class imbalance can lead to model selection bias toward the majority class. To overcome this problem, therefore, more class-oriented metrics can be formulated, such as \textit{sensitivity} (TPR), \text{specificity} (TNR), and \text{precision} (PPV).

\begin{equation}
\label{eq:base}
TPR = \frac{TP}{TP + FN}, \quad TNR= \frac{TN}{TN + FP}, \quad PPV = \frac{TP}{TP+FP}.
\end{equation}

These metrics are based solely on the values of the confusion matrix and can often be referred to as \textit{base metrics}. Of course, the premise of classification is to identify all instances as best as possible, so a presentation of all of them is required to get a complete picture of how a classifier works. However, many research areas show a tendency to express quality by single metrics. In the case of imbalanced data classification, several metrics have been developed that aggregate the values of the underlying metrics. These metrics (hereafter referred to as aggregated metrics) often use averaging of \textit{base metrics}, for example: \textit{balanced accuracy} (BAC), $G_{mean}$ or $F_{\beta}$ are \textit{arithmetic mean}, \textit{geometric mean} and \textit{harmonic mean}, respectively (equation \ref{eq:agg})~\cite{Krawczyk:2016}.

%These metrics are based solely on the values of the confusion matrix and can often be referred to as \textit{base metrics}. Of course, the premise of the classification is to recognize all instances as best as possible, so base metrics have to be combined in specific ways to achieve this goal. The aggregation methods often employ an averaging of base metrics, for instance: \textit{balanced accuracy} (BAC), $G_{mean}$ or $F_{\beta}$ is the \textit{arithmetic mean}, \textit{geometric mean} and \textit{harmonic mean}, respectively (Equation \ref{eq:agg}).

\begin{align}
\label{eq:agg}
\begin{split}
BAC = &\frac{TPR + TNR}{2} \quad G_{mean} = \sqrt{TPR \times TNR} \quad \\
\vspace{0.5em} \\
& F_{\beta} = \frac{(\beta^2+1) \times {PPV} \times {TPR} } {\beta^2 \times {PPV} + {TPR} }
\end{split}
\end{align}

As mentioned in the introduction, aggregated metrics lead to ambiguities, such as multiple pairs of (TPR,~TNR) for which the metric has the same value. Therefore, it is essential to analyze the metrics not just by a single value but rather by their underlying components, although this is often a neglected step in research.

\subsection{Multi-objective optimization}

%On the other hand, there are existing methods for solving problems using multiple criteria. 
In general, a \textit{multi-objective optimization} (\textsc{moo}) task can be described as:

\begin{equation}
\textit{maximize} \quad f_i(x) \qquad i = 1, 2, ... M\; \\
\end{equation}

\noindent where $f(x)$ represents a set of criterion values to evaluate a solution $x$, creating the $M$-dimensional space of objective functions.

In the case of \textsc{moo}, the multiplicity of criteria indicates that not one solution but a set of solutions is returned. The criteria describing such solutions are often connected, and therefore increasing one criterion may decrease another. The optimal set of such solutions forms a \textit{Pareto front}. The algorithm's objective is to create a set of approximate solutions that are as close as possible in terms of their optimality to the \textit{Pareto front}. 

Let us denote $P$ as a \textit{Pareto front approximation} discovered by the algorithm, where $p \in P$ is a single instance of the solution and $r$ is a reference solution. % to compare with. 
It is necessary to determine their relationship in the space of objective functions to establish a hierarchy between p and r. We will denote \textit{strict dominance} (SD)  as

\begin{align*}
SD: p \succ r \quad \iff \quad \forall_{i=1}^{M} f_i(p) > f_i(r) \\
\end{align*}

This relationship helps to choose better solutions. % and reject the worst. 
Of course, beyond such comparison, there can be solutions that neither dominate nor are dominated. For such a reason \textsc{moo} solutions must be evaluated as a complete front. Many metrics allow you to compare non-dominant solutions returned by MOO algorithms~\cite{laszczyk2019}. For such a reason, e.g.,  the \textit{Hypervolume} (HV) metric is proposed

\begin{equation}
HV (P, r) = \lambda ( \bigcup_{p \in P}[p, r])
\end{equation}

\noindent where $\lambda$ is the Lebesgue measure. The provided formula describes the field determined by the relative solutions concerning a reference point $r$. Consequently, it accounts for both the distance between $P$ and $r$ and the width of the front.

Another frequently used metric is \textit{Generational Distance} (GD):

\begin{equation}
GD (P;R) = \frac{1}{||P||}(\sum_{p \in P} \min_{r \in R} || p - r ||)
\end{equation}

The value of GD is calculated from the set of reference solutions, defined as $R$. In the specific case where $|| R || = 1$, this metric is also referred to as the \textit{Euclidean Distance} (ED). It can be interpreted as the average distance between the solutions and the reference point.

\section{Proposed Methodology}

Despite the natural possibility of employing \textsc{moo} metrics to assess \textsc{mol} performance, their use focuses primarily on evaluating the attributes of the generated \textit{Pareto front approximation} rather than analyzing the quality of individual classification criteria, which in the context of the imbalanced data problem, these are $TPR$ and $TNR$. Although calculating an aggregate metric for each solution is possible, this approach may result in the loss of information regarding the diversity of solutions due to the ambiguity of the metrics mentioned above.

A fair \textsc{mol} algorithm comparison can only be accomplished by establishing a reference solution. %Moreover, 
It is essential to emphasize that such a comparison should prioritize estimating the diversity and dominance of the solutions available in \textsc{mol} instead of a direct comparison. From a classification perspective, quantitative factors may be relevant in characterizing the relationship between the front and the reference classifier regarding the dominance of solutions within the criterion function space. To this end, we propose two metrics, the Strict Dominance Ratio ($SDR$) and the Non-dominated Ratio ($NDR$), which are formulated as follows:

\begin{equation}
SDR = \frac{|| P \succ r||}{|| P ||}, \quad NDR = 1 - \frac{|| P \prec r ||}{|| P ||}
\end{equation}

These metrics determine the percentage of \textsc{mol} solutions that improve the reference algorithm in considered criteria. While the $SDR$ is more desirable from the standpoint of answering the question "Is a given algorithm better?" it is also essential to consider $NDR$ as the degree to which solutions improve the diversity of possible solutions.

\begin{figure}
\centering
\begin{subfigure}{0.45\linewidth}
\includegraphics[width=\linewidth]{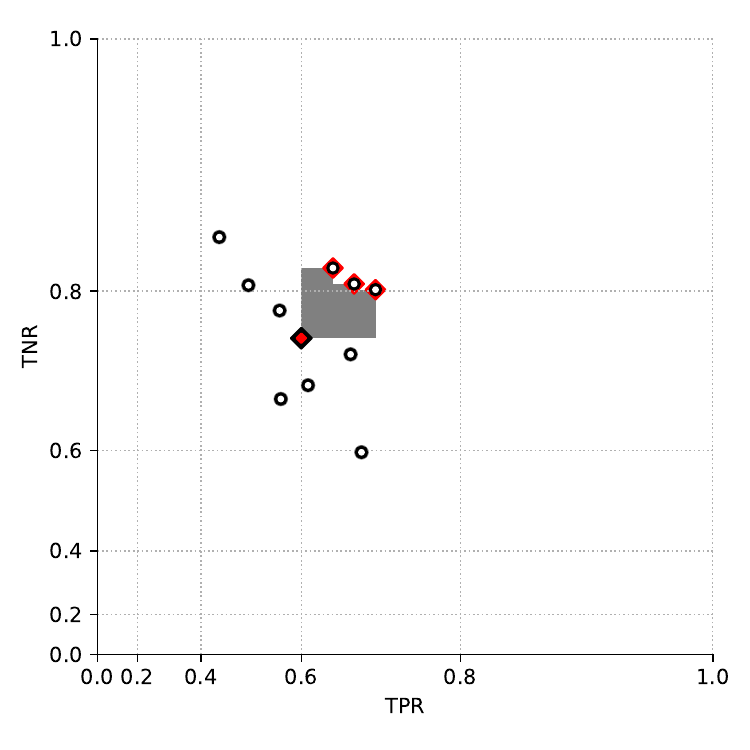} 
\caption{Hypervolume}
\end{subfigure}\hfill
\begin{subfigure}{0.45\linewidth}
\includegraphics[width=\linewidth]{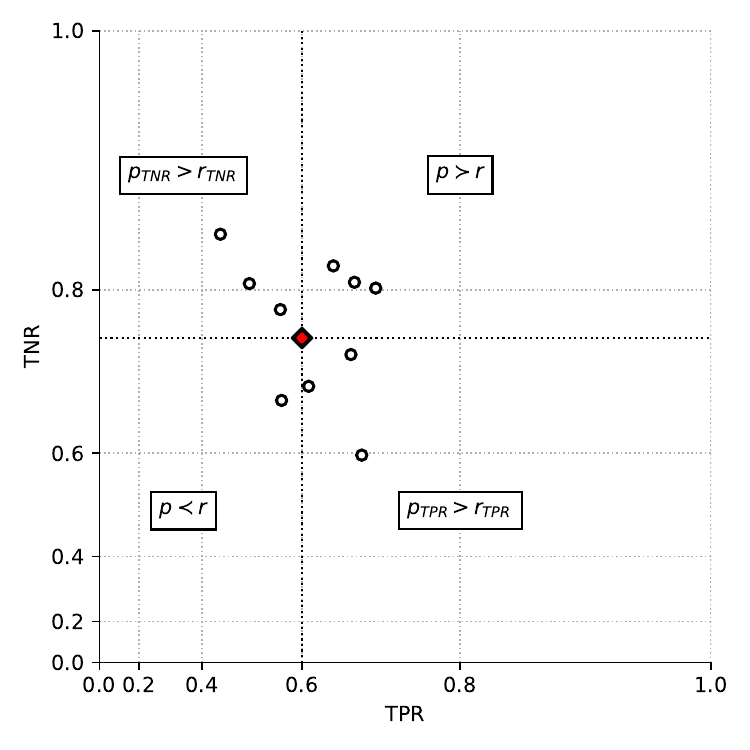} 
\caption{Dominance regions}
\end{subfigure}\hfill
\caption{Comparison of graphical representation of metrics.}
\label{fig:compare}
\end{figure}

An illustrative comparison of the $HV$ area with dominance regions is presented in Figure \ref{fig:compare}.

%Although direct use of metrics for aggregation has been avoided, one of them, the parametric measure $F_{\beta}$, can be used to some extent. Because the $\beta$ parameter controls the user preference, it can be used to establish a connection with the preference of the \textsc{mol} solutions. 
Although direct use of aggregated metrics has been avoided, one of them, the parametric measure $F_{\beta}$, can be used to some extent. Because the $\beta$ parameter controls the user preference (i.e., how much more important is $recall$ than $precision$), it can be used to establish a connection with the preference of the \textsc{mol} solutions. The $F_{\beta}$-plot can be employed \cite{wojciechowski2024} to perform such a comparison. Due to the extensive range of solutions, only a version without a statistic test (for a single split) is feasible. By plotting the curve for both the reference algorithms and the \textsc{mol} algorithm, it is possible to identify the area of user preference in which the evaluated algorithm has the potential to enhance the results.

\section {Case Study}

To introduce the proposed evaluation methodology, a series of experiments were conducted using a resampling technique based on \textsc{moo}, such as Multi-objective Evolutionary Undersampling (MEUS) ~\cite{wojciechowski2021}.

Although the research questions that guide the experiment are important, the case study's primary objective remains the evaluation of the proposed methods in terms of their ability to provide a fair assessment of the \textsc{mol} model compared to the reference algorithms.

\subsection{Assumptions}
The experiments mainly investigate two key questions: 
\begin{itemize}
\item whether the \textsc{moo} based algorithm may be compared to single solutions (\textbf{Experiment~A}), 
\item whether the investigated algorithm can improve the quality of recognition when applied to a combined set of reference models (\textbf{Experiment~B}).
\end{itemize}

To carry out the experiments, 10 datasets were selected from the \textit{keel} repository \cite{derrac2015} that exhibited varying degrees of \textit{imbalance ratio} (IR) (between ~$2$ and $82$). These datasets were chosen on the basis of their diversity and to ensure that the minority class size would be sufficient to guarantee a representative subset for use in the model evaluation. The selected datasets and their characteristics are presented in Table \ref{tab:datasets}.

\begin{table}
\caption{Selected datasets characteristics.}
\label{tab:datasets}
\begin{tabularx}{\columnwidth}{llRRRR}
\toprule
$\mathcal{DS}_{n} $ & $ \mathcal{DS}_{name} $                &   $\mathcal{F}$ &   $|\mathcal{DS}|$ &   $|\mathcal{DS}_{min}|$ &       $IR$ \\
\midrule
$\mathcal{DS}_{1} $ & \texttt{pima}                       &            8    &         768        &                      268 &  1.87      \\
$\mathcal{DS}_{2} $ & \texttt{vehicle1}                   &           18    &         846        &                      217 &  2.90      \\
$\mathcal{DS}_{3} $ & \texttt{new-thyroid1}               &            5    &         215        &                       35 &  5.14      \\
$\mathcal{DS}_{4} $ & \texttt{segment0}                   &           19    &        2308        &                      329 &  6.02      \\
$\mathcal{DS}_{5} $ & \texttt{page-blocks0}               &           10    &        5472        &                      559 &  8.79      \\
$\mathcal{DS}_{6} $ & \texttt{yeast-0-2-5-6\_vs\_3-7-8-9} &            8    &        1004        &                       99 &  9.14      \\
$\mathcal{DS}_{7} $ & \texttt{shuttle-c0-vs-c4}           &            9    &        1829        &                      123 & 13.87      \\
$\mathcal{DS}_{8} $ & \texttt{ecoli4}                     &            7    &         336        &                       20 & 15.8       \\
$\mathcal{DS}_{9} $ & \texttt{winequality-white-3\_vs\_7} &           11    &         900        &                       20 & 44.00      \\
$\mathcal{DS}_{10} $ & \texttt{poker-8-9\_vs\_5}          &           10    &        2075        &                       25 & 82.00      \\
\bottomrule
\end{tabularx}
\end{table}

A $5\times2$ \textit{cross-validation} with stratified data split was employed to stabilize the experimental results. In \textbf{A}, the results were averaged across all splits, whereas, due to the limitations of \textbf{B}, the results are only presented for a single split (equivalent to a 50:50 split).

The reference pool was created by combining 10 oversampling and 10 undersampling algorithms with the base model, which was \textit{Gaussian Naive Bayes} (GNB). In addition, a model without preprocessing (NoSMOTE) was included in the pool. The parameters of the algorithms were set to their standard configuration, while the entire environment configuration and the scripts to reproduce the experiment can be found in the publicly available repository \footnote{https://anonymous.4open.science/r/moo-compare-ECML25/} % \footnote{https://github.com/w4k2/moo-compare}.

\subsection{Experiment A: Results}
The results of MEUS comparison with reference algorithms were assessed using \textsc{moo} metrics (Table \ref{tab:moo}) and the proposed set of metrics (Table \ref{tab:mol}). The standard deviation of the metric is provided under each score.
 
In the case of the \textit{GD}, it is noteworthy that for the sets $\mathcal{DS}_4$, $\mathcal{DS}_9$, and $\mathcal{DS}_{10}$, the distance to most reference solutions is either close to or equal to $0$. While there are exceptions, it can be stated with a reasonable degree of certainty that for the sets above, MEUS cannot introduce solutions significantly different from those offered by the selected resamplings. It is also possible to observe high scores, particularly in comparison with NearMiss. However, in the case of \textit{GD}, this distance may also be attributed to the aforementioned algorithm's inadequate adaptation to those datasets. Nevertheless, it is noteworthy that for the sets $\mathcal{DS}_1$, $\mathcal{DS}_2$, $\mathcal{DS}_3$, and $\mathcal{DS}_6$, the \textit{GD} value is never $0$, indicating that the algorithm can often introduce a diverse range of solutions. The conclusions above are also reflected in the values of the \textit{HV} metric, where results approaching 0 are observed exclusively for the $\mathcal{DS}_4$ set.

\begin{table}[]
\caption{Results for \textsc{moo} metrics.}
\label{tab:moo}
\begin{adjustbox}{totalheight=\textheight-2\baselineskip,center}
\scriptsize{\begin{tabularx}{\columnwidth}{Lcccccccccc}
\toprule
 Sampling                  & $\mathcal{DS}_{1}$                               & $\mathcal{DS}_{2}$                               & $\mathcal{DS}_{3}$                              & $\mathcal{DS}_{4}$                             & $\mathcal{DS}_{5}$                               & $\mathcal{DS}_{6}$                             & $\mathcal{DS}_{7}$                               & $\mathcal{DS}_{8}$                             & $\mathcal{DS}_{9}$                             & $\mathcal{DS}_{10}$                            \\
\midrule
\multicolumn{11}{c}{ED}\\
\midrule
 NoSMOTE                   & \makecell{2.74 \\ \tiny{ \color{gray} (2.69)}}   & \makecell{1.67 \\ \tiny{ \color{gray} (0.99)}}   & \makecell{0.25 \\ \tiny{ \color{gray} (0.29)}}  & \makecell{0.00 \\ \tiny{ \color{gray} (0.00)}} & \makecell{0.44 \\ \tiny{ \color{gray} (0.98)}}   & \makecell{0.18 \\ \tiny{ \color{gray} (0.37)}} & \makecell{9.09 \\ \tiny{ \color{gray} (2.28)}}   & \makecell{0.00 \\ \tiny{ \color{gray} (0.00)}} & \makecell{0.00 \\ \tiny{ \color{gray} (0.00)}} & \makecell{0.00 \\ \tiny{ \color{gray} (0.00)}} \\
 RandomOverSampler         & \makecell{1.04 \\ \tiny{ \color{gray} (1.11)}}   & \makecell{1.71 \\ \tiny{ \color{gray} (0.94)}}   & \makecell{0.38 \\ \tiny{ \color{gray} (0.48)}}  & \makecell{0.00 \\ \tiny{ \color{gray} (0.00)}} & \makecell{0.00 \\ \tiny{ \color{gray} (0.00)}}   & \makecell{0.09 \\ \tiny{ \color{gray} (0.14)}} & \makecell{10.47 \\ \tiny{ \color{gray} (1.74)}}  & \makecell{0.00 \\ \tiny{ \color{gray} (0.00)}} & \makecell{0.00 \\ \tiny{ \color{gray} (0.00)}} & \makecell{0.00 \\ \tiny{ \color{gray} (0.00)}} \\
 SMOTE                     & \makecell{1.81 \\ \tiny{ \color{gray} (1.83)}}   & \makecell{1.72 \\ \tiny{ \color{gray} (0.91)}}   & \makecell{1.49 \\ \tiny{ \color{gray} (2.51)}}  & \makecell{0.00 \\ \tiny{ \color{gray} (0.00)}} & \makecell{0.00 \\ \tiny{ \color{gray} (0.00)}}   & \makecell{0.12 \\ \tiny{ \color{gray} (0.24)}} & \makecell{9.05 \\ \tiny{ \color{gray} (2.42)}}   & \makecell{1.09 \\ \tiny{ \color{gray} (3.27)}} & \makecell{0.00 \\ \tiny{ \color{gray} (0.00)}} & \makecell{0.00 \\ \tiny{ \color{gray} (0.00)}} \\
 ProWSyn                   & \makecell{2.65 \\ \tiny{ \color{gray} (2.95)}}   & \makecell{1.55 \\ \tiny{ \color{gray} (0.97)}}   & \makecell{0.50 \\ \tiny{ \color{gray} (0.69)}}  & \makecell{0.00 \\ \tiny{ \color{gray} (0.00)}} & \makecell{0.00 \\ \tiny{ \color{gray} (0.00)}}   & \makecell{0.11 \\ \tiny{ \color{gray} (0.18)}} & \makecell{8.26 \\ \tiny{ \color{gray} (2.20)}}   & \makecell{0.73 \\ \tiny{ \color{gray} (1.47)}} & \makecell{0.00 \\ \tiny{ \color{gray} (0.00)}} & \makecell{0.00 \\ \tiny{ \color{gray} (0.00)}} \\
 BorderlineSMOTE           & \makecell{0.43 \\ \tiny{ \color{gray} (0.48)}}   & \makecell{26.38 \\ \tiny{ \color{gray} (29.41)}} & \makecell{1.22 \\ \tiny{ \color{gray} (1.47)}}  & \makecell{0.00 \\ \tiny{ \color{gray} (0.00)}} & \makecell{3.10 \\ \tiny{ \color{gray} (7.28)}}   & \makecell{0.05 \\ \tiny{ \color{gray} (0.14)}} & \makecell{6.49 \\ \tiny{ \color{gray} (2.66)}}   & \makecell{4.23 \\ \tiny{ \color{gray} (5.54)}} & \makecell{0.00 \\ \tiny{ \color{gray} (0.00)}} & \makecell{3.32 \\ \tiny{ \color{gray} (6.84)}} \\
 DBSMOTE                   & \makecell{4.95 \\ \tiny{ \color{gray} (4.55)}}   & \makecell{2.05 \\ \tiny{ \color{gray} (0.97)}}   & \makecell{7.03 \\ \tiny{ \color{gray} (11.97)}} & \makecell{0.00 \\ \tiny{ \color{gray} (0.00)}} & \makecell{0.95 \\ \tiny{ \color{gray} (1.95)}}   & \makecell{1.38 \\ \tiny{ \color{gray} (1.44)}} & \makecell{0.74 \\ \tiny{ \color{gray} (0.55)}}   & \makecell{0.00 \\ \tiny{ \color{gray} (0.00)}} & \makecell{0.65 \\ \tiny{ \color{gray} (1.09)}} & \makecell{0.21 \\ \tiny{ \color{gray} (0.32)}} \\
 SOMO                      & \makecell{12.69 \\ \tiny{ \color{gray} (24.78)}} & \makecell{9.88 \\ \tiny{ \color{gray} (7.92)}}   & \makecell{0.36 \\ \tiny{ \color{gray} (0.71)}}  & \makecell{0.01 \\ \tiny{ \color{gray} (0.02)}} & \makecell{0.25 \\ \tiny{ \color{gray} (0.58)}}   & \makecell{0.12 \\ \tiny{ \color{gray} (0.15)}} & \makecell{10.78 \\ \tiny{ \color{gray} (3.07)}}  & \makecell{1.86 \\ \tiny{ \color{gray} (5.37)}} & \makecell{0.00 \\ \tiny{ \color{gray} (0.00)}} & \makecell{0.71 \\ \tiny{ \color{gray} (2.14)}} \\
 MSYN                      & \makecell{3.50 \\ \tiny{ \color{gray} (2.99)}}   & \makecell{2.84 \\ \tiny{ \color{gray} (1.49)}}   & \makecell{1.50 \\ \tiny{ \color{gray} (3.07)}}  & \makecell{0.00 \\ \tiny{ \color{gray} (0.00)}} & \makecell{0.25 \\ \tiny{ \color{gray} (0.51)}}   & \makecell{0.05 \\ \tiny{ \color{gray} (0.13)}} & \makecell{9.65 \\ \tiny{ \color{gray} (2.32)}}   & \makecell{3.91 \\ \tiny{ \color{gray} (6.19)}} & \makecell{0.00 \\ \tiny{ \color{gray} (0.00)}} & \makecell{0.00 \\ \tiny{ \color{gray} (0.00)}} \\
 CCR                       & \makecell{2.95 \\ \tiny{ \color{gray} (2.42)}}   & \makecell{1.76 \\ \tiny{ \color{gray} (1.02)}}   & \makecell{0.37 \\ \tiny{ \color{gray} (0.75)}}  & \makecell{0.00 \\ \tiny{ \color{gray} (0.00)}} & \makecell{7.47 \\ \tiny{ \color{gray} (10.96)}}  & \makecell{0.07 \\ \tiny{ \color{gray} (0.11)}} & \makecell{8.85 \\ \tiny{ \color{gray} (1.74)}}   & \makecell{0.00 \\ \tiny{ \color{gray} (0.00)}} & \makecell{0.00 \\ \tiny{ \color{gray} (0.00)}} & \makecell{0.00 \\ \tiny{ \color{gray} (0.00)}} \\
 AHC                       & \makecell{1.22 \\ \tiny{ \color{gray} (1.53)}}   & \makecell{1.98 \\ \tiny{ \color{gray} (0.96)}}   & \makecell{0.19 \\ \tiny{ \color{gray} (0.18)}}  & \makecell{0.00 \\ \tiny{ \color{gray} (0.00)}} & \makecell{0.63 \\ \tiny{ \color{gray} (1.52)}}   & \makecell{0.04 \\ \tiny{ \color{gray} (0.11)}} & \makecell{10.04 \\ \tiny{ \color{gray} (1.40)}}  & \makecell{0.25 \\ \tiny{ \color{gray} (0.56)}} & \makecell{0.00 \\ \tiny{ \color{gray} (0.00)}} & \makecell{0.23 \\ \tiny{ \color{gray} (0.46)}} \\
 ADASYN                    & \makecell{0.70 \\ \tiny{ \color{gray} (0.77)}}   & \makecell{1.27 \\ \tiny{ \color{gray} (1.64)}}   & \makecell{0.91 \\ \tiny{ \color{gray} (0.93)}}  & \makecell{0.00 \\ \tiny{ \color{gray} (0.01)}} & \makecell{0.00 \\ \tiny{ \color{gray} (0.00)}}   & \makecell{0.01 \\ \tiny{ \color{gray} (0.03)}} & \makecell{5.42 \\ \tiny{ \color{gray} (2.46)}}   & \makecell{1.41 \\ \tiny{ \color{gray} (4.23)}} & \makecell{0.00 \\ \tiny{ \color{gray} (0.00)}} & \makecell{0.00 \\ \tiny{ \color{gray} (0.00)}} \\
 RandomUnderSampler        & \makecell{3.38 \\ \tiny{ \color{gray} (4.35)}}   & \makecell{1.79 \\ \tiny{ \color{gray} (0.71)}}   & \makecell{0.72 \\ \tiny{ \color{gray} (1.49)}}  & \makecell{0.00 \\ \tiny{ \color{gray} (0.00)}} & \makecell{3.80 \\ \tiny{ \color{gray} (5.94)}}   & \makecell{0.15 \\ \tiny{ \color{gray} (0.34)}} & \makecell{8.63 \\ \tiny{ \color{gray} (3.37)}}   & \makecell{0.00 \\ \tiny{ \color{gray} (0.00)}} & \makecell{0.00 \\ \tiny{ \color{gray} (0.00)}} & \makecell{0.00 \\ \tiny{ \color{gray} (0.00)}} \\
 ClusterCentroids          & \makecell{1.13 \\ \tiny{ \color{gray} (1.59)}}   & \makecell{2.25 \\ \tiny{ \color{gray} (1.16)}}   & \makecell{2.07 \\ \tiny{ \color{gray} (2.60)}}  & \makecell{0.00 \\ \tiny{ \color{gray} (0.01)}} & \makecell{0.00 \\ \tiny{ \color{gray} (0.00)}}   & \makecell{0.03 \\ \tiny{ \color{gray} (0.09)}} & \makecell{0.00 \\ \tiny{ \color{gray} (0.00)}}   & \makecell{0.00 \\ \tiny{ \color{gray} (0.00)}} & \makecell{0.00 \\ \tiny{ \color{gray} (0.00)}} & \makecell{0.00 \\ \tiny{ \color{gray} (0.00)}} \\
 InstanceHardnessThreshold & \makecell{1.30 \\ \tiny{ \color{gray} (1.29)}}   & \makecell{1.81 \\ \tiny{ \color{gray} (0.97)}}   & \makecell{1.56 \\ \tiny{ \color{gray} (2.02)}}  & \makecell{0.00 \\ \tiny{ \color{gray} (0.00)}} & \makecell{4.11 \\ \tiny{ \color{gray} (7.74)}}   & \makecell{0.01 \\ \tiny{ \color{gray} (0.04)}} & \makecell{0.91 \\ \tiny{ \color{gray} (0.89)}}   & \makecell{0.84 \\ \tiny{ \color{gray} (1.70)}} & \makecell{0.00 \\ \tiny{ \color{gray} (0.00)}} & \makecell{1.50 \\ \tiny{ \color{gray} (4.49)}} \\
 NearMiss                  & \makecell{5.83 \\ \tiny{ \color{gray} (4.52)}}   & \makecell{73.51 \\ \tiny{ \color{gray} (16.79)}} & \makecell{47.54 \\ \tiny{ \color{gray} (9.36)}} & \makecell{0.00 \\ \tiny{ \color{gray} (0.00)}} & \makecell{13.73 \\ \tiny{ \color{gray} (19.42)}} & \makecell{1.19 \\ \tiny{ \color{gray} (1.55)}} & \makecell{78.93 \\ \tiny{ \color{gray} (15.71)}} & \makecell{0.00 \\ \tiny{ \color{gray} (0.00)}} & \makecell{0.00 \\ \tiny{ \color{gray} (0.00)}} & \makecell{0.00 \\ \tiny{ \color{gray} (0.00)}} \\
 TomekLinks                & \makecell{2.13 \\ \tiny{ \color{gray} (2.18)}}   & \makecell{1.72 \\ \tiny{ \color{gray} (0.93)}}   & \makecell{0.19 \\ \tiny{ \color{gray} (0.23)}}  & \makecell{0.00 \\ \tiny{ \color{gray} (0.00)}} & \makecell{0.44 \\ \tiny{ \color{gray} (0.98)}}   & \makecell{0.08 \\ \tiny{ \color{gray} (0.17)}} & \makecell{9.23 \\ \tiny{ \color{gray} (1.94)}}   & \makecell{0.00 \\ \tiny{ \color{gray} (0.00)}} & \makecell{0.00 \\ \tiny{ \color{gray} (0.00)}} & \makecell{0.00 \\ \tiny{ \color{gray} (0.00)}} \\
 EditedNearestNeighbours   & \makecell{1.95 \\ \tiny{ \color{gray} (1.76)}}   & \makecell{1.69 \\ \tiny{ \color{gray} (0.99)}}   & \makecell{0.37 \\ \tiny{ \color{gray} (0.52)}}  & \makecell{0.00 \\ \tiny{ \color{gray} (0.00)}} & \makecell{7.03 \\ \tiny{ \color{gray} (13.59)}}  & \makecell{0.07 \\ \tiny{ \color{gray} (0.13)}} & \makecell{7.92 \\ \tiny{ \color{gray} (2.01)}}   & \makecell{0.00 \\ \tiny{ \color{gray} (0.00)}} & \makecell{0.00 \\ \tiny{ \color{gray} (0.00)}} & \makecell{0.00 \\ \tiny{ \color{gray} (0.00)}} \\
 AllKNN                    & \makecell{2.01 \\ \tiny{ \color{gray} (1.70)}}   & \makecell{1.69 \\ \tiny{ \color{gray} (1.00)}}   & \makecell{0.35 \\ \tiny{ \color{gray} (0.47)}}  & \makecell{0.00 \\ \tiny{ \color{gray} (0.00)}} & \makecell{11.96 \\ \tiny{ \color{gray} (14.95)}} & \makecell{0.01 \\ \tiny{ \color{gray} (0.04)}} & \makecell{6.82 \\ \tiny{ \color{gray} (1.92)}}   & \makecell{0.00 \\ \tiny{ \color{gray} (0.00)}} & \makecell{0.00 \\ \tiny{ \color{gray} (0.00)}} & \makecell{0.00 \\ \tiny{ \color{gray} (0.00)}} \\
 OneSidedSelection         & \makecell{1.70 \\ \tiny{ \color{gray} (1.61)}}   & \makecell{1.99 \\ \tiny{ \color{gray} (1.13)}}   & \makecell{0.51 \\ \tiny{ \color{gray} (0.98)}}  & \makecell{1.07 \\ \tiny{ \color{gray} (3.21)}} & \makecell{0.51 \\ \tiny{ \color{gray} (1.52)}}   & \makecell{0.07 \\ \tiny{ \color{gray} (0.12)}} & \makecell{5.01 \\ \tiny{ \color{gray} (1.85)}}   & \makecell{0.00 \\ \tiny{ \color{gray} (0.00)}} & \makecell{0.00 \\ \tiny{ \color{gray} (0.00)}} & \makecell{0.00 \\ \tiny{ \color{gray} (0.00)}} \\
 CondensedNearestNeighbour & \makecell{1.70 \\ \tiny{ \color{gray} (1.22)}}   & \makecell{1.92 \\ \tiny{ \color{gray} (0.99)}}   & \makecell{2.40 \\ \tiny{ \color{gray} (3.01)}}  & \makecell{1.07 \\ \tiny{ \color{gray} (3.21)}} & \makecell{0.00 \\ \tiny{ \color{gray} (0.00)}}   & \makecell{0.08 \\ \tiny{ \color{gray} (0.11)}} & \makecell{9.74 \\ \tiny{ \color{gray} (4.01)}}   & \makecell{0.45 \\ \tiny{ \color{gray} (1.36)}} & \makecell{0.83 \\ \tiny{ \color{gray} (1.67)}} & \makecell{0.18 \\ \tiny{ \color{gray} (0.37)}} \\
 NeighbourhoodCleaningRule & \makecell{1.42 \\ \tiny{ \color{gray} (1.53)}}   & \makecell{1.67 \\ \tiny{ \color{gray} (0.99)}}   & \makecell{0.19 \\ \tiny{ \color{gray} (0.23)}}  & \makecell{0.00 \\ \tiny{ \color{gray} (0.00)}} & \makecell{7.03 \\ \tiny{ \color{gray} (13.59)}}  & \makecell{0.03 \\ \tiny{ \color{gray} (0.05)}} & \makecell{7.99 \\ \tiny{ \color{gray} (2.32)}}   & \makecell{0.02 \\ \tiny{ \color{gray} (0.07)}} & \makecell{0.00 \\ \tiny{ \color{gray} (0.00)}} & \makecell{0.00 \\ \tiny{ \color{gray} (0.00)}} \\
\midrule
\multicolumn{11}{c}{HV ($\times 10^{3}$)}\\
\midrule
 NoSMOTE                   & \makecell{0.16 \\ \tiny{ \color{gray} (0.03)}} & \makecell{0.16 \\ \tiny{ \color{gray} (0.04)}} & \makecell{0.27 \\ \tiny{ \color{gray} (0.15)}} & \makecell{0.01 \\ \tiny{ \color{gray} (0.01)}} & \makecell{0.23 \\ \tiny{ \color{gray} (0.11)}} & \makecell{0.23 \\ \tiny{ \color{gray} (0.02)}} & \makecell{0.25 \\ \tiny{ \color{gray} (0.02)}} & \makecell{0.23 \\ \tiny{ \color{gray} (0.09)}} & \makecell{0.09 \\ \tiny{ \color{gray} (0.03)}} & \makecell{0.04 \\ \tiny{ \color{gray} (0.03)}} \\
 RandomOverSampler         & \makecell{0.20 \\ \tiny{ \color{gray} (0.04)}} & \makecell{0.18 \\ \tiny{ \color{gray} (0.04)}} & \makecell{0.23 \\ \tiny{ \color{gray} (0.16)}} & \makecell{0.00 \\ \tiny{ \color{gray} (0.00)}} & \makecell{0.27 \\ \tiny{ \color{gray} (0.13)}} & \makecell{0.23 \\ \tiny{ \color{gray} (0.03)}} & \makecell{0.20 \\ \tiny{ \color{gray} (0.03)}} & \makecell{0.37 \\ \tiny{ \color{gray} (0.13)}} & \makecell{0.11 \\ \tiny{ \color{gray} (0.03)}} & \makecell{0.46 \\ \tiny{ \color{gray} (0.09)}} \\
 SMOTE                     & \makecell{0.20 \\ \tiny{ \color{gray} (0.03)}} & \makecell{0.17 \\ \tiny{ \color{gray} (0.04)}} & \makecell{0.23 \\ \tiny{ \color{gray} (0.15)}} & \makecell{0.00 \\ \tiny{ \color{gray} (0.01)}} & \makecell{0.19 \\ \tiny{ \color{gray} (0.07)}} & \makecell{0.23 \\ \tiny{ \color{gray} (0.02)}} & \makecell{0.16 \\ \tiny{ \color{gray} (0.04)}} & \makecell{0.36 \\ \tiny{ \color{gray} (0.17)}} & \makecell{0.10 \\ \tiny{ \color{gray} (0.03)}} & \makecell{0.40 \\ \tiny{ \color{gray} (0.11)}} \\
 ProWSyn                   & \makecell{0.18 \\ \tiny{ \color{gray} (0.03)}} & \makecell{0.10 \\ \tiny{ \color{gray} (0.04)}} & \makecell{0.23 \\ \tiny{ \color{gray} (0.14)}} & \makecell{0.00 \\ \tiny{ \color{gray} (0.01)}} & \makecell{0.18 \\ \tiny{ \color{gray} (0.07)}} & \makecell{0.23 \\ \tiny{ \color{gray} (0.02)}} & \makecell{0.16 \\ \tiny{ \color{gray} (0.05)}} & \makecell{0.29 \\ \tiny{ \color{gray} (0.13)}} & \makecell{0.10 \\ \tiny{ \color{gray} (0.03)}} & \makecell{0.38 \\ \tiny{ \color{gray} (0.08)}} \\
 BorderlineSMOTE           & \makecell{0.30 \\ \tiny{ \color{gray} (0.08)}} & \makecell{0.25 \\ \tiny{ \color{gray} (0.09)}} & \makecell{0.24 \\ \tiny{ \color{gray} (0.15)}} & \makecell{0.01 \\ \tiny{ \color{gray} (0.01)}} & \makecell{0.19 \\ \tiny{ \color{gray} (0.07)}} & \makecell{0.24 \\ \tiny{ \color{gray} (0.03)}} & \makecell{0.10 \\ \tiny{ \color{gray} (0.03)}} & \makecell{0.25 \\ \tiny{ \color{gray} (0.09)}} & \makecell{0.11 \\ \tiny{ \color{gray} (0.04)}} & \makecell{0.22 \\ \tiny{ \color{gray} (0.10)}} \\
 DBSMOTE                   & \makecell{0.19 \\ \tiny{ \color{gray} (0.04)}} & \makecell{0.19 \\ \tiny{ \color{gray} (0.04)}} & \makecell{0.25 \\ \tiny{ \color{gray} (0.13)}} & \makecell{0.00 \\ \tiny{ \color{gray} (0.01)}} & \makecell{0.24 \\ \tiny{ \color{gray} (0.13)}} & \makecell{0.24 \\ \tiny{ \color{gray} (0.02)}} & \makecell{0.05 \\ \tiny{ \color{gray} (0.01)}} & \makecell{0.37 \\ \tiny{ \color{gray} (0.13)}} & \makecell{0.10 \\ \tiny{ \color{gray} (0.04)}} & \makecell{0.25 \\ \tiny{ \color{gray} (0.22)}} \\
 SOMO                      & \makecell{0.20 \\ \tiny{ \color{gray} (0.06)}} & \makecell{0.25 \\ \tiny{ \color{gray} (0.15)}} & \makecell{0.28 \\ \tiny{ \color{gray} (0.12)}} & \makecell{0.01 \\ \tiny{ \color{gray} (0.01)}} & \makecell{0.15 \\ \tiny{ \color{gray} (0.06)}} & \makecell{0.23 \\ \tiny{ \color{gray} (0.02)}} & \makecell{0.29 \\ \tiny{ \color{gray} (0.03)}} & \makecell{0.23 \\ \tiny{ \color{gray} (0.09)}} & \makecell{0.09 \\ \tiny{ \color{gray} (0.04)}} & \makecell{0.04 \\ \tiny{ \color{gray} (0.04)}} \\
 MSYN                      & \makecell{0.17 \\ \tiny{ \color{gray} (0.02)}} & \makecell{0.17 \\ \tiny{ \color{gray} (0.04)}} & \makecell{0.20 \\ \tiny{ \color{gray} (0.08)}} & \makecell{0.00 \\ \tiny{ \color{gray} (0.01)}} & \makecell{0.17 \\ \tiny{ \color{gray} (0.05)}} & \makecell{0.23 \\ \tiny{ \color{gray} (0.02)}} & \makecell{0.17 \\ \tiny{ \color{gray} (0.04)}} & \makecell{0.32 \\ \tiny{ \color{gray} (0.12)}} & \makecell{0.10 \\ \tiny{ \color{gray} (0.03)}} & \makecell{0.49 \\ \tiny{ \color{gray} (0.12)}} \\
 CCR                       & \makecell{0.19 \\ \tiny{ \color{gray} (0.04)}} & \makecell{0.09 \\ \tiny{ \color{gray} (0.04)}} & \makecell{0.20 \\ \tiny{ \color{gray} (0.07)}} & \makecell{0.01 \\ \tiny{ \color{gray} (0.01)}} & \makecell{0.40 \\ \tiny{ \color{gray} (0.25)}} & \makecell{0.23 \\ \tiny{ \color{gray} (0.02)}} & \makecell{0.16 \\ \tiny{ \color{gray} (0.03)}} & \makecell{0.39 \\ \tiny{ \color{gray} (0.14)}} & \makecell{0.11 \\ \tiny{ \color{gray} (0.03)}} & \makecell{0.46 \\ \tiny{ \color{gray} (0.10)}} \\
 AHC                       & \makecell{0.18 \\ \tiny{ \color{gray} (0.04)}} & \makecell{0.16 \\ \tiny{ \color{gray} (0.04)}} & \makecell{0.23 \\ \tiny{ \color{gray} (0.14)}} & \makecell{0.00 \\ \tiny{ \color{gray} (0.01)}} & \makecell{0.19 \\ \tiny{ \color{gray} (0.08)}} & \makecell{0.24 \\ \tiny{ \color{gray} (0.02)}} & \makecell{0.22 \\ \tiny{ \color{gray} (0.02)}} & \makecell{0.22 \\ \tiny{ \color{gray} (0.09)}} & \makecell{0.10 \\ \tiny{ \color{gray} (0.03)}} & \makecell{0.05 \\ \tiny{ \color{gray} (0.03)}} \\
 ADASYN                    & \makecell{0.26 \\ \tiny{ \color{gray} (0.06)}} & \makecell{0.20 \\ \tiny{ \color{gray} (0.04)}} & \makecell{0.24 \\ \tiny{ \color{gray} (0.15)}} & \makecell{0.01 \\ \tiny{ \color{gray} (0.02)}} & \makecell{0.20 \\ \tiny{ \color{gray} (0.07)}} & \makecell{0.24 \\ \tiny{ \color{gray} (0.02)}} & \makecell{0.09 \\ \tiny{ \color{gray} (0.03)}} & \makecell{0.36 \\ \tiny{ \color{gray} (0.16)}} & \makecell{0.11 \\ \tiny{ \color{gray} (0.04)}} & \makecell{0.41 \\ \tiny{ \color{gray} (0.11)}} \\
 RandomUnderSampler        & \makecell{0.20 \\ \tiny{ \color{gray} (0.04)}} & \makecell{0.17 \\ \tiny{ \color{gray} (0.04)}} & \makecell{0.35 \\ \tiny{ \color{gray} (0.13)}} & \makecell{0.01 \\ \tiny{ \color{gray} (0.02)}} & \makecell{0.33 \\ \tiny{ \color{gray} (0.22)}} & \makecell{0.23 \\ \tiny{ \color{gray} (0.03)}} & \makecell{0.14 \\ \tiny{ \color{gray} (0.05)}} & \makecell{0.57 \\ \tiny{ \color{gray} (0.19)}} & \makecell{0.11 \\ \tiny{ \color{gray} (0.03)}} & \makecell{0.67 \\ \tiny{ \color{gray} (0.11)}} \\
 ClusterCentroids          & \makecell{0.30 \\ \tiny{ \color{gray} (0.07)}} & \makecell{0.23 \\ \tiny{ \color{gray} (0.03)}} & \makecell{0.27 \\ \tiny{ \color{gray} (0.15)}} & \makecell{0.00 \\ \tiny{ \color{gray} (0.00)}} & \makecell{0.29 \\ \tiny{ \color{gray} (0.15)}} & \makecell{0.23 \\ \tiny{ \color{gray} (0.02)}} & \makecell{0.36 \\ \tiny{ \color{gray} (0.03)}} & \makecell{0.54 \\ \tiny{ \color{gray} (0.10)}} & \makecell{0.14 \\ \tiny{ \color{gray} (0.03)}} & \makecell{0.88 \\ \tiny{ \color{gray} (0.19)}} \\
 InstanceHardnessThreshold & \makecell{0.30 \\ \tiny{ \color{gray} (0.07)}} & \makecell{0.18 \\ \tiny{ \color{gray} (0.04)}} & \makecell{0.36 \\ \tiny{ \color{gray} (0.11)}} & \makecell{0.01 \\ \tiny{ \color{gray} (0.01)}} & \makecell{0.27 \\ \tiny{ \color{gray} (0.15)}} & \makecell{0.31 \\ \tiny{ \color{gray} (0.03)}} & \makecell{0.10 \\ \tiny{ \color{gray} (0.02)}} & \makecell{0.29 \\ \tiny{ \color{gray} (0.10)}} & \makecell{0.10 \\ \tiny{ \color{gray} (0.03)}} & \makecell{0.07 \\ \tiny{ \color{gray} (0.06)}} \\
 NearMiss                  & \makecell{0.17 \\ \tiny{ \color{gray} (0.03)}} & \makecell{0.34 \\ \tiny{ \color{gray} (0.05)}} & \makecell{0.31 \\ \tiny{ \color{gray} (0.05)}} & \makecell{0.02 \\ \tiny{ \color{gray} (0.02)}} & \makecell{0.47 \\ \tiny{ \color{gray} (0.24)}} & \makecell{0.26 \\ \tiny{ \color{gray} (0.02)}} & \makecell{0.46 \\ \tiny{ \color{gray} (0.04)}} & \makecell{1.02 \\ \tiny{ \color{gray} (0.14)}} & \makecell{0.08 \\ \tiny{ \color{gray} (0.03)}} & \makecell{1.14 \\ \tiny{ \color{gray} (0.09)}} \\
 TomekLinks                & \makecell{0.17 \\ \tiny{ \color{gray} (0.03)}} & \makecell{0.16 \\ \tiny{ \color{gray} (0.04)}} & \makecell{0.30 \\ \tiny{ \color{gray} (0.18)}} & \makecell{0.01 \\ \tiny{ \color{gray} (0.01)}} & \makecell{0.23 \\ \tiny{ \color{gray} (0.11)}} & \makecell{0.23 \\ \tiny{ \color{gray} (0.02)}} & \makecell{0.23 \\ \tiny{ \color{gray} (0.02)}} & \makecell{0.23 \\ \tiny{ \color{gray} (0.09)}} & \makecell{0.09 \\ \tiny{ \color{gray} (0.03)}} & \makecell{0.04 \\ \tiny{ \color{gray} (0.03)}} \\
 EditedNearestNeighbours   & \makecell{0.20 \\ \tiny{ \color{gray} (0.04)}} & \makecell{0.17 \\ \tiny{ \color{gray} (0.04)}} & \makecell{0.38 \\ \tiny{ \color{gray} (0.11)}} & \makecell{0.01 \\ \tiny{ \color{gray} (0.01)}} & \makecell{0.38 \\ \tiny{ \color{gray} (0.25)}} & \makecell{0.28 \\ \tiny{ \color{gray} (0.03)}} & \makecell{0.13 \\ \tiny{ \color{gray} (0.03)}} & \makecell{0.23 \\ \tiny{ \color{gray} (0.09)}} & \makecell{0.10 \\ \tiny{ \color{gray} (0.03)}} & \makecell{0.04 \\ \tiny{ \color{gray} (0.03)}} \\
 AllKNN                    & \makecell{0.21 \\ \tiny{ \color{gray} (0.05)}} & \makecell{0.17 \\ \tiny{ \color{gray} (0.04)}} & \makecell{0.38 \\ \tiny{ \color{gray} (0.11)}} & \makecell{0.01 \\ \tiny{ \color{gray} (0.01)}} & \makecell{0.53 \\ \tiny{ \color{gray} (0.29)}} & \makecell{0.32 \\ \tiny{ \color{gray} (0.03)}} & \makecell{0.11 \\ \tiny{ \color{gray} (0.03)}} & \makecell{0.23 \\ \tiny{ \color{gray} (0.09)}} & \makecell{0.10 \\ \tiny{ \color{gray} (0.03)}} & \makecell{0.04 \\ \tiny{ \color{gray} (0.03)}} \\
 OneSidedSelection         & \makecell{0.17 \\ \tiny{ \color{gray} (0.03)}} & \makecell{0.17 \\ \tiny{ \color{gray} (0.04)}} & \makecell{0.30 \\ \tiny{ \color{gray} (0.18)}} & \makecell{0.28 \\ \tiny{ \color{gray} (0.24)}} & \makecell{0.26 \\ \tiny{ \color{gray} (0.14)}} & \makecell{0.23 \\ \tiny{ \color{gray} (0.02)}} & \makecell{0.26 \\ \tiny{ \color{gray} (0.03)}} & \makecell{0.24 \\ \tiny{ \color{gray} (0.09)}} & \makecell{0.08 \\ \tiny{ \color{gray} (0.02)}} & \makecell{0.04 \\ \tiny{ \color{gray} (0.03)}} \\
 CondensedNearestNeighbour & \makecell{0.19 \\ \tiny{ \color{gray} (0.03)}} & \makecell{0.11 \\ \tiny{ \color{gray} (0.04)}} & \makecell{0.22 \\ \tiny{ \color{gray} (0.06)}} & \makecell{0.44 \\ \tiny{ \color{gray} (0.29)}} & \makecell{0.21 \\ \tiny{ \color{gray} (0.09)}} & \makecell{0.24 \\ \tiny{ \color{gray} (0.02)}} & \makecell{0.41 \\ \tiny{ \color{gray} (0.03)}} & \makecell{0.34 \\ \tiny{ \color{gray} (0.16)}} & \makecell{0.08 \\ \tiny{ \color{gray} (0.03)}} & \makecell{0.06 \\ \tiny{ \color{gray} (0.05)}} \\
 NeighbourhoodCleaningRule & \makecell{0.19 \\ \tiny{ \color{gray} (0.04)}} & \makecell{0.16 \\ \tiny{ \color{gray} (0.04)}} & \makecell{0.41 \\ \tiny{ \color{gray} (0.14)}} & \makecell{0.01 \\ \tiny{ \color{gray} (0.01)}} & \makecell{0.38 \\ \tiny{ \color{gray} (0.25)}} & \makecell{0.25 \\ \tiny{ \color{gray} (0.02)}} & \makecell{0.14 \\ \tiny{ \color{gray} (0.04)}} & \makecell{0.23 \\ \tiny{ \color{gray} (0.09)}} & \makecell{0.09 \\ \tiny{ \color{gray} (0.03)}} & \makecell{0.04 \\ \tiny{ \color{gray} (0.03)}} \\
\bottomrule
\end{tabularx}}
\end{adjustbox}
\end{table}

\begin{table}[]
\caption{Results for proposed metrics.}
\label{tab:mol}
\begin{adjustbox}{totalheight=\textheight-2\baselineskip,center}
\scriptsize{\begin{tabularx}{\columnwidth}{Lcccccccccc}
\toprule
 Sampling                  & $\mathcal{DS}_{1}$                             & $\mathcal{DS}_{2}$                             & $\mathcal{DS}_{3}$                             & $\mathcal{DS}_{4}$                             & $\mathcal{DS}_{5}$                             & $\mathcal{DS}_{6}$                             & $\mathcal{DS}_{7}$                             & $\mathcal{DS}_{8}$                             & $\mathcal{DS}_{9}$                             & $\mathcal{DS}_{10}$                            \\
\midrule
\multicolumn{11}{c}{SRD}\\
\midrule
 NoSMOTE                   & \makecell{0.20 \\ \tiny{ \color{gray} (0.15)}} & \makecell{0.30 \\ \tiny{ \color{gray} (0.20)}} & \makecell{0.07 \\ \tiny{ \color{gray} (0.07)}} & \makecell{0.00 \\ \tiny{ \color{gray} (0.00)}} & \makecell{0.06 \\ \tiny{ \color{gray} (0.12)}} & \makecell{0.03 \\ \tiny{ \color{gray} (0.04)}} & \makecell{0.98 \\ \tiny{ \color{gray} (0.05)}} & \makecell{0.00 \\ \tiny{ \color{gray} (0.00)}} & \makecell{0.00 \\ \tiny{ \color{gray} (0.00)}} & \makecell{0.00 \\ \tiny{ \color{gray} (0.00)}} \\
 RandomOverSampler         & \makecell{0.11 \\ \tiny{ \color{gray} (0.12)}} & \makecell{0.29 \\ \tiny{ \color{gray} (0.19)}} & \makecell{0.11 \\ \tiny{ \color{gray} (0.10)}} & \makecell{0.00 \\ \tiny{ \color{gray} (0.00)}} & \makecell{0.00 \\ \tiny{ \color{gray} (0.00)}} & \makecell{0.02 \\ \tiny{ \color{gray} (0.02)}} & \makecell{1.00 \\ \tiny{ \color{gray} (0.00)}} & \makecell{0.00 \\ \tiny{ \color{gray} (0.00)}} & \makecell{0.00 \\ \tiny{ \color{gray} (0.00)}} & \makecell{0.00 \\ \tiny{ \color{gray} (0.00)}} \\
 SMOTE                     & \makecell{0.14 \\ \tiny{ \color{gray} (0.13)}} & \makecell{0.30 \\ \tiny{ \color{gray} (0.19)}} & \makecell{0.13 \\ \tiny{ \color{gray} (0.12)}} & \makecell{0.00 \\ \tiny{ \color{gray} (0.00)}} & \makecell{0.00 \\ \tiny{ \color{gray} (0.00)}} & \makecell{0.02 \\ \tiny{ \color{gray} (0.03)}} & \makecell{0.99 \\ \tiny{ \color{gray} (0.03)}} & \makecell{0.01 \\ \tiny{ \color{gray} (0.03)}} & \makecell{0.00 \\ \tiny{ \color{gray} (0.00)}} & \makecell{0.00 \\ \tiny{ \color{gray} (0.00)}} \\
 ProWSyn                   & \makecell{0.18 \\ \tiny{ \color{gray} (0.17)}} & \makecell{0.33 \\ \tiny{ \color{gray} (0.17)}} & \makecell{0.10 \\ \tiny{ \color{gray} (0.11)}} & \makecell{0.00 \\ \tiny{ \color{gray} (0.00)}} & \makecell{0.00 \\ \tiny{ \color{gray} (0.00)}} & \makecell{0.02 \\ \tiny{ \color{gray} (0.02)}} & \makecell{0.97 \\ \tiny{ \color{gray} (0.05)}} & \makecell{0.11 \\ \tiny{ \color{gray} (0.21)}} & \makecell{0.00 \\ \tiny{ \color{gray} (0.00)}} & \makecell{0.00 \\ \tiny{ \color{gray} (0.00)}} \\
 BorderlineSMOTE           & \makecell{0.05 \\ \tiny{ \color{gray} (0.05)}} & \makecell{0.67 \\ \tiny{ \color{gray} (0.30)}} & \makecell{0.15 \\ \tiny{ \color{gray} (0.14)}} & \makecell{0.00 \\ \tiny{ \color{gray} (0.00)}} & \makecell{0.14 \\ \tiny{ \color{gray} (0.23)}} & \makecell{0.01 \\ \tiny{ \color{gray} (0.03)}} & \makecell{0.83 \\ \tiny{ \color{gray} (0.15)}} & \makecell{0.23 \\ \tiny{ \color{gray} (0.25)}} & \makecell{0.00 \\ \tiny{ \color{gray} (0.00)}} & \makecell{0.10 \\ \tiny{ \color{gray} (0.27)}} \\
 DBSMOTE                   & \makecell{0.25 \\ \tiny{ \color{gray} (0.19)}} & \makecell{0.31 \\ \tiny{ \color{gray} (0.18)}} & \makecell{0.26 \\ \tiny{ \color{gray} (0.19)}} & \makecell{0.00 \\ \tiny{ \color{gray} (0.00)}} & \makecell{0.07 \\ \tiny{ \color{gray} (0.12)}} & \makecell{0.09 \\ \tiny{ \color{gray} (0.07)}} & \makecell{0.40 \\ \tiny{ \color{gray} (0.19)}} & \makecell{0.00 \\ \tiny{ \color{gray} (0.00)}} & \makecell{0.10 \\ \tiny{ \color{gray} (0.16)}} & \makecell{0.09 \\ \tiny{ \color{gray} (0.14)}} \\
 SOMO                      & \makecell{0.23 \\ \tiny{ \color{gray} (0.27)}} & \makecell{0.53 \\ \tiny{ \color{gray} (0.17)}} & \makecell{0.08 \\ \tiny{ \color{gray} (0.10)}} & \makecell{0.10 \\ \tiny{ \color{gray} (0.30)}} & \makecell{0.01 \\ \tiny{ \color{gray} (0.02)}} & \makecell{0.03 \\ \tiny{ \color{gray} (0.04)}} & \makecell{0.97 \\ \tiny{ \color{gray} (0.05)}} & \makecell{0.10 \\ \tiny{ \color{gray} (0.23)}} & \makecell{0.00 \\ \tiny{ \color{gray} (0.00)}} & \makecell{0.01 \\ \tiny{ \color{gray} (0.01)}} \\
 MSYN                      & \makecell{0.20 \\ \tiny{ \color{gray} (0.16)}} & \makecell{0.42 \\ \tiny{ \color{gray} (0.20)}} & \makecell{0.12 \\ \tiny{ \color{gray} (0.12)}} & \makecell{0.00 \\ \tiny{ \color{gray} (0.00)}} & \makecell{0.03 \\ \tiny{ \color{gray} (0.08)}} & \makecell{0.02 \\ \tiny{ \color{gray} (0.03)}} & \makecell{0.99 \\ \tiny{ \color{gray} (0.03)}} & \makecell{0.14 \\ \tiny{ \color{gray} (0.24)}} & \makecell{0.00 \\ \tiny{ \color{gray} (0.00)}} & \makecell{0.00 \\ \tiny{ \color{gray} (0.00)}} \\
 CCR                       & \makecell{0.21 \\ \tiny{ \color{gray} (0.14)}} & \makecell{0.40 \\ \tiny{ \color{gray} (0.23)}} & \makecell{0.07 \\ \tiny{ \color{gray} (0.07)}} & \makecell{0.00 \\ \tiny{ \color{gray} (0.00)}} & \makecell{0.23 \\ \tiny{ \color{gray} (0.26)}} & \makecell{0.02 \\ \tiny{ \color{gray} (0.03)}} & \makecell{1.00 \\ \tiny{ \color{gray} (0.00)}} & \makecell{0.00 \\ \tiny{ \color{gray} (0.00)}} & \makecell{0.00 \\ \tiny{ \color{gray} (0.00)}} & \makecell{0.00 \\ \tiny{ \color{gray} (0.00)}} \\
 AHC                       & \makecell{0.12 \\ \tiny{ \color{gray} (0.13)}} & \makecell{0.33 \\ \tiny{ \color{gray} (0.18)}} & \makecell{0.08 \\ \tiny{ \color{gray} (0.08)}} & \makecell{0.00 \\ \tiny{ \color{gray} (0.00)}} & \makecell{0.03 \\ \tiny{ \color{gray} (0.06)}} & \makecell{0.01 \\ \tiny{ \color{gray} (0.03)}} & \makecell{1.00 \\ \tiny{ \color{gray} (0.00)}} & \makecell{0.03 \\ \tiny{ \color{gray} (0.05)}} & \makecell{0.00 \\ \tiny{ \color{gray} (0.00)}} & \makecell{0.10 \\ \tiny{ \color{gray} (0.27)}} \\
 ADASYN                    & \makecell{0.08 \\ \tiny{ \color{gray} (0.06)}} & \makecell{0.20 \\ \tiny{ \color{gray} (0.23)}} & \makecell{0.15 \\ \tiny{ \color{gray} (0.12)}} & \makecell{0.10 \\ \tiny{ \color{gray} (0.30)}} & \makecell{0.00 \\ \tiny{ \color{gray} (0.00)}} & \makecell{0.01 \\ \tiny{ \color{gray} (0.01)}} & \makecell{0.81 \\ \tiny{ \color{gray} (0.14)}} & \makecell{0.01 \\ \tiny{ \color{gray} (0.04)}} & \makecell{0.00 \\ \tiny{ \color{gray} (0.00)}} & \makecell{0.00 \\ \tiny{ \color{gray} (0.00)}} \\
 RandomUnderSampler        & \makecell{0.20 \\ \tiny{ \color{gray} (0.17)}} & \makecell{0.33 \\ \tiny{ \color{gray} (0.15)}} & \makecell{0.11 \\ \tiny{ \color{gray} (0.16)}} & \makecell{0.00 \\ \tiny{ \color{gray} (0.00)}} & \makecell{0.16 \\ \tiny{ \color{gray} (0.26)}} & \makecell{0.02 \\ \tiny{ \color{gray} (0.05)}} & \makecell{0.93 \\ \tiny{ \color{gray} (0.13)}} & \makecell{0.00 \\ \tiny{ \color{gray} (0.00)}} & \makecell{0.00 \\ \tiny{ \color{gray} (0.00)}} & \makecell{0.00 \\ \tiny{ \color{gray} (0.00)}} \\
 ClusterCentroids          & \makecell{0.06 \\ \tiny{ \color{gray} (0.05)}} & \makecell{0.29 \\ \tiny{ \color{gray} (0.19)}} & \makecell{0.15 \\ \tiny{ \color{gray} (0.15)}} & \makecell{0.01 \\ \tiny{ \color{gray} (0.04)}} & \makecell{0.00 \\ \tiny{ \color{gray} (0.00)}} & \makecell{0.01 \\ \tiny{ \color{gray} (0.01)}} & \makecell{0.00 \\ \tiny{ \color{gray} (0.00)}} & \makecell{0.00 \\ \tiny{ \color{gray} (0.00)}} & \makecell{0.00 \\ \tiny{ \color{gray} (0.00)}} & \makecell{0.00 \\ \tiny{ \color{gray} (0.00)}} \\
 InstanceHardnessThreshold & \makecell{0.08 \\ \tiny{ \color{gray} (0.07)}} & \makecell{0.30 \\ \tiny{ \color{gray} (0.19)}} & \makecell{0.14 \\ \tiny{ \color{gray} (0.15)}} & \makecell{0.00 \\ \tiny{ \color{gray} (0.00)}} & \makecell{0.18 \\ \tiny{ \color{gray} (0.31)}} & \makecell{0.01 \\ \tiny{ \color{gray} (0.01)}} & \makecell{0.12 \\ \tiny{ \color{gray} (0.08)}} & \makecell{0.02 \\ \tiny{ \color{gray} (0.04)}} & \makecell{0.00 \\ \tiny{ \color{gray} (0.00)}} & \makecell{0.11 \\ \tiny{ \color{gray} (0.27)}} \\
 NearMiss                  & \makecell{0.27 \\ \tiny{ \color{gray} (0.15)}} & \makecell{0.94 \\ \tiny{ \color{gray} (0.11)}} & \makecell{0.59 \\ \tiny{ \color{gray} (0.16)}} & \makecell{0.00 \\ \tiny{ \color{gray} (0.00)}} & \makecell{0.33 \\ \tiny{ \color{gray} (0.41)}} & \makecell{0.06 \\ \tiny{ \color{gray} (0.05)}} & \makecell{0.93 \\ \tiny{ \color{gray} (0.08)}} & \makecell{0.00 \\ \tiny{ \color{gray} (0.00)}} & \makecell{0.00 \\ \tiny{ \color{gray} (0.00)}} & \makecell{0.00 \\ \tiny{ \color{gray} (0.00)}} \\
 TomekLinks                & \makecell{0.17 \\ \tiny{ \color{gray} (0.14)}} & \makecell{0.31 \\ \tiny{ \color{gray} (0.19)}} & \makecell{0.06 \\ \tiny{ \color{gray} (0.07)}} & \makecell{0.00 \\ \tiny{ \color{gray} (0.00)}} & \makecell{0.06 \\ \tiny{ \color{gray} (0.12)}} & \makecell{0.01 \\ \tiny{ \color{gray} (0.02)}} & \makecell{0.98 \\ \tiny{ \color{gray} (0.05)}} & \makecell{0.00 \\ \tiny{ \color{gray} (0.00)}} & \makecell{0.00 \\ \tiny{ \color{gray} (0.00)}} & \makecell{0.00 \\ \tiny{ \color{gray} (0.00)}} \\
 EditedNearestNeighbours   & \makecell{0.14 \\ \tiny{ \color{gray} (0.13)}} & \makecell{0.30 \\ \tiny{ \color{gray} (0.20)}} & \makecell{0.08 \\ \tiny{ \color{gray} (0.08)}} & \makecell{0.00 \\ \tiny{ \color{gray} (0.00)}} & \makecell{0.16 \\ \tiny{ \color{gray} (0.27)}} & \makecell{0.01 \\ \tiny{ \color{gray} (0.02)}} & \makecell{0.97 \\ \tiny{ \color{gray} (0.05)}} & \makecell{0.00 \\ \tiny{ \color{gray} (0.00)}} & \makecell{0.00 \\ \tiny{ \color{gray} (0.00)}} & \makecell{0.00 \\ \tiny{ \color{gray} (0.00)}} \\
 AllKNN                    & \makecell{0.14 \\ \tiny{ \color{gray} (0.10)}} & \makecell{0.30 \\ \tiny{ \color{gray} (0.20)}} & \makecell{0.08 \\ \tiny{ \color{gray} (0.08)}} & \makecell{0.00 \\ \tiny{ \color{gray} (0.00)}} & \makecell{0.25 \\ \tiny{ \color{gray} (0.32)}} & \makecell{0.01 \\ \tiny{ \color{gray} (0.02)}} & \makecell{0.89 \\ \tiny{ \color{gray} (0.10)}} & \makecell{0.00 \\ \tiny{ \color{gray} (0.00)}} & \makecell{0.00 \\ \tiny{ \color{gray} (0.00)}} & \makecell{0.00 \\ \tiny{ \color{gray} (0.00)}} \\
 OneSidedSelection         & \makecell{0.16 \\ \tiny{ \color{gray} (0.11)}} & \makecell{0.34 \\ \tiny{ \color{gray} (0.16)}} & \makecell{0.10 \\ \tiny{ \color{gray} (0.16)}} & \makecell{0.01 \\ \tiny{ \color{gray} (0.04)}} & \makecell{0.06 \\ \tiny{ \color{gray} (0.18)}} & \makecell{0.02 \\ \tiny{ \color{gray} (0.03)}} & \makecell{0.81 \\ \tiny{ \color{gray} (0.12)}} & \makecell{0.00 \\ \tiny{ \color{gray} (0.00)}} & \makecell{0.00 \\ \tiny{ \color{gray} (0.00)}} & \makecell{0.00 \\ \tiny{ \color{gray} (0.00)}} \\
 CondensedNearestNeighbour & \makecell{0.14 \\ \tiny{ \color{gray} (0.10)}} & \makecell{0.41 \\ \tiny{ \color{gray} (0.19)}} & \makecell{0.18 \\ \tiny{ \color{gray} (0.20)}} & \makecell{0.01 \\ \tiny{ \color{gray} (0.04)}} & \makecell{0.00 \\ \tiny{ \color{gray} (0.00)}} & \makecell{0.03 \\ \tiny{ \color{gray} (0.03)}} & \makecell{0.84 \\ \tiny{ \color{gray} (0.20)}} & \makecell{0.01 \\ \tiny{ \color{gray} (0.03)}} & \makecell{0.07 \\ \tiny{ \color{gray} (0.15)}} & \makecell{0.03 \\ \tiny{ \color{gray} (0.06)}} \\
 NeighbourhoodCleaningRule & \makecell{0.14 \\ \tiny{ \color{gray} (0.12)}} & \makecell{0.30 \\ \tiny{ \color{gray} (0.20)}} & \makecell{0.06 \\ \tiny{ \color{gray} (0.08)}} & \makecell{0.00 \\ \tiny{ \color{gray} (0.00)}} & \makecell{0.16 \\ \tiny{ \color{gray} (0.27)}} & \makecell{0.02 \\ \tiny{ \color{gray} (0.02)}} & \makecell{0.93 \\ \tiny{ \color{gray} (0.08)}} & \makecell{0.01 \\ \tiny{ \color{gray} (0.02)}} & \makecell{0.00 \\ \tiny{ \color{gray} (0.00)}} & \makecell{0.00 \\ \tiny{ \color{gray} (0.00)}} \\
\midrule
\multicolumn{11}{c}{NDR}\\
\midrule
 NoSMOTE                   & \makecell{0.99 \\ \tiny{ \color{gray} (0.03)}} & \makecell{1.00 \\ \tiny{ \color{gray} (0.00)}} & \makecell{0.99 \\ \tiny{ \color{gray} (0.02)}} & \makecell{0.29 \\ \tiny{ \color{gray} (0.43)}} & \makecell{0.94 \\ \tiny{ \color{gray} (0.12)}} & \makecell{0.98 \\ \tiny{ \color{gray} (0.03)}} & \makecell{1.00 \\ \tiny{ \color{gray} (0.00)}} & \makecell{0.78 \\ \tiny{ \color{gray} (0.12)}} & \makecell{0.78 \\ \tiny{ \color{gray} (0.28)}} & \makecell{0.37 \\ \tiny{ \color{gray} (0.32)}} \\
 RandomOverSampler         & \makecell{1.00 \\ \tiny{ \color{gray} (0.00)}} & \makecell{1.00 \\ \tiny{ \color{gray} (0.00)}} & \makecell{0.98 \\ \tiny{ \color{gray} (0.03)}} & \makecell{0.32 \\ \tiny{ \color{gray} (0.45)}} & \makecell{0.94 \\ \tiny{ \color{gray} (0.09)}} & \makecell{0.97 \\ \tiny{ \color{gray} (0.05)}} & \makecell{1.00 \\ \tiny{ \color{gray} (0.00)}} & \makecell{0.99 \\ \tiny{ \color{gray} (0.03)}} & \makecell{0.99 \\ \tiny{ \color{gray} (0.03)}} & \makecell{1.00 \\ \tiny{ \color{gray} (0.00)}} \\
 SMOTE                     & \makecell{0.99 \\ \tiny{ \color{gray} (0.02)}} & \makecell{1.00 \\ \tiny{ \color{gray} (0.00)}} & \makecell{0.99 \\ \tiny{ \color{gray} (0.02)}} & \makecell{0.23 \\ \tiny{ \color{gray} (0.39)}} & \makecell{0.89 \\ \tiny{ \color{gray} (0.11)}} & \makecell{0.96 \\ \tiny{ \color{gray} (0.04)}} & \makecell{1.00 \\ \tiny{ \color{gray} (0.00)}} & \makecell{0.98 \\ \tiny{ \color{gray} (0.03)}} & \makecell{0.91 \\ \tiny{ \color{gray} (0.12)}} & \makecell{1.00 \\ \tiny{ \color{gray} (0.00)}} \\
 ProWSyn                   & \makecell{0.99 \\ \tiny{ \color{gray} (0.03)}} & \makecell{1.00 \\ \tiny{ \color{gray} (0.00)}} & \makecell{0.99 \\ \tiny{ \color{gray} (0.02)}} & \makecell{0.12 \\ \tiny{ \color{gray} (0.30)}} & \makecell{0.86 \\ \tiny{ \color{gray} (0.14)}} & \makecell{0.96 \\ \tiny{ \color{gray} (0.04)}} & \makecell{1.00 \\ \tiny{ \color{gray} (0.00)}} & \makecell{0.97 \\ \tiny{ \color{gray} (0.05)}} & \makecell{0.91 \\ \tiny{ \color{gray} (0.12)}} & \makecell{1.00 \\ \tiny{ \color{gray} (0.00)}} \\
 BorderlineSMOTE           & \makecell{1.00 \\ \tiny{ \color{gray} (0.00)}} & \makecell{1.00 \\ \tiny{ \color{gray} (0.00)}} & \makecell{0.96 \\ \tiny{ \color{gray} (0.08)}} & \makecell{0.29 \\ \tiny{ \color{gray} (0.43)}} & \makecell{0.93 \\ \tiny{ \color{gray} (0.07)}} & \makecell{0.97 \\ \tiny{ \color{gray} (0.03)}} & \makecell{1.00 \\ \tiny{ \color{gray} (0.00)}} & \makecell{0.98 \\ \tiny{ \color{gray} (0.05)}} & \makecell{0.92 \\ \tiny{ \color{gray} (0.13)}} & \makecell{0.98 \\ \tiny{ \color{gray} (0.03)}} \\
 DBSMOTE                   & \makecell{0.99 \\ \tiny{ \color{gray} (0.02)}} & \makecell{1.00 \\ \tiny{ \color{gray} (0.00)}} & \makecell{1.00 \\ \tiny{ \color{gray} (0.00)}} & \makecell{0.12 \\ \tiny{ \color{gray} (0.30)}} & \makecell{0.79 \\ \tiny{ \color{gray} (0.30)}} & \makecell{0.99 \\ \tiny{ \color{gray} (0.02)}} & \makecell{1.00 \\ \tiny{ \color{gray} (0.00)}} & \makecell{0.99 \\ \tiny{ \color{gray} (0.03)}} & \makecell{0.89 \\ \tiny{ \color{gray} (0.30)}} & \makecell{0.97 \\ \tiny{ \color{gray} (0.08)}} \\
 SOMO                      & \makecell{1.00 \\ \tiny{ \color{gray} (0.00)}} & \makecell{1.00 \\ \tiny{ \color{gray} (0.00)}} & \makecell{0.96 \\ \tiny{ \color{gray} (0.12)}} & \makecell{0.89 \\ \tiny{ \color{gray} (0.28)}} & \makecell{0.80 \\ \tiny{ \color{gray} (0.17)}} & \makecell{0.94 \\ \tiny{ \color{gray} (0.07)}} & \makecell{1.00 \\ \tiny{ \color{gray} (0.00)}} & \makecell{0.78 \\ \tiny{ \color{gray} (0.16)}} & \makecell{0.73 \\ \tiny{ \color{gray} (0.33)}} & \makecell{0.40 \\ \tiny{ \color{gray} (0.38)}} \\
 MSYN                      & \makecell{0.99 \\ \tiny{ \color{gray} (0.02)}} & \makecell{1.00 \\ \tiny{ \color{gray} (0.00)}} & \makecell{0.98 \\ \tiny{ \color{gray} (0.03)}} & \makecell{0.23 \\ \tiny{ \color{gray} (0.39)}} & \makecell{0.80 \\ \tiny{ \color{gray} (0.23)}} & \makecell{0.96 \\ \tiny{ \color{gray} (0.04)}} & \makecell{1.00 \\ \tiny{ \color{gray} (0.00)}} & \makecell{0.98 \\ \tiny{ \color{gray} (0.05)}} & \makecell{0.80 \\ \tiny{ \color{gray} (0.29)}} & \makecell{1.00 \\ \tiny{ \color{gray} (0.00)}} \\
 CCR                       & \makecell{1.00 \\ \tiny{ \color{gray} (0.00)}} & \makecell{1.00 \\ \tiny{ \color{gray} (0.00)}} & \makecell{0.97 \\ \tiny{ \color{gray} (0.03)}} & \makecell{0.80 \\ \tiny{ \color{gray} (0.40)}} & \makecell{0.91 \\ \tiny{ \color{gray} (0.12)}} & \makecell{0.97 \\ \tiny{ \color{gray} (0.04)}} & \makecell{1.00 \\ \tiny{ \color{gray} (0.00)}} & \makecell{0.99 \\ \tiny{ \color{gray} (0.03)}} & \makecell{0.97 \\ \tiny{ \color{gray} (0.05)}} & \makecell{1.00 \\ \tiny{ \color{gray} (0.00)}} \\
 AHC                       & \makecell{0.99 \\ \tiny{ \color{gray} (0.02)}} & \makecell{1.00 \\ \tiny{ \color{gray} (0.00)}} & \makecell{0.98 \\ \tiny{ \color{gray} (0.03)}} & \makecell{0.23 \\ \tiny{ \color{gray} (0.39)}} & \makecell{0.93 \\ \tiny{ \color{gray} (0.09)}} & \makecell{0.96 \\ \tiny{ \color{gray} (0.04)}} & \makecell{1.00 \\ \tiny{ \color{gray} (0.00)}} & \makecell{0.85 \\ \tiny{ \color{gray} (0.09)}} & \makecell{0.78 \\ \tiny{ \color{gray} (0.28)}} & \makecell{0.52 \\ \tiny{ \color{gray} (0.41)}} \\
 ADASYN                    & \makecell{0.99 \\ \tiny{ \color{gray} (0.02)}} & \makecell{0.99 \\ \tiny{ \color{gray} (0.03)}} & \makecell{0.99 \\ \tiny{ \color{gray} (0.02)}} & \makecell{0.36 \\ \tiny{ \color{gray} (0.44)}} & \makecell{0.93 \\ \tiny{ \color{gray} (0.08)}} & \makecell{0.97 \\ \tiny{ \color{gray} (0.05)}} & \makecell{1.00 \\ \tiny{ \color{gray} (0.00)}} & \makecell{0.99 \\ \tiny{ \color{gray} (0.03)}} & \makecell{0.92 \\ \tiny{ \color{gray} (0.13)}} & \makecell{1.00 \\ \tiny{ \color{gray} (0.00)}} \\
 RandomUnderSampler        & \makecell{0.99 \\ \tiny{ \color{gray} (0.02)}} & \makecell{1.00 \\ \tiny{ \color{gray} (0.00)}} & \makecell{0.97 \\ \tiny{ \color{gray} (0.05)}} & \makecell{0.78 \\ \tiny{ \color{gray} (0.40)}} & \makecell{0.84 \\ \tiny{ \color{gray} (0.19)}} & \makecell{0.96 \\ \tiny{ \color{gray} (0.05)}} & \makecell{1.00 \\ \tiny{ \color{gray} (0.00)}} & \makecell{1.00 \\ \tiny{ \color{gray} (0.00)}} & \makecell{0.97 \\ \tiny{ \color{gray} (0.05)}} & \makecell{1.00 \\ \tiny{ \color{gray} (0.00)}} \\
 ClusterCentroids          & \makecell{0.99 \\ \tiny{ \color{gray} (0.02)}} & \makecell{1.00 \\ \tiny{ \color{gray} (0.00)}} & \makecell{0.98 \\ \tiny{ \color{gray} (0.02)}} & \makecell{0.51 \\ \tiny{ \color{gray} (0.50)}} & \makecell{0.92 \\ \tiny{ \color{gray} (0.12)}} & \makecell{0.96 \\ \tiny{ \color{gray} (0.06)}} & \makecell{1.00 \\ \tiny{ \color{gray} (0.00)}} & \makecell{1.00 \\ \tiny{ \color{gray} (0.00)}} & \makecell{1.00 \\ \tiny{ \color{gray} (0.00)}} & \makecell{1.00 \\ \tiny{ \color{gray} (0.00)}} \\
 InstanceHardnessThreshold & \makecell{1.00 \\ \tiny{ \color{gray} (0.00)}} & \makecell{1.00 \\ \tiny{ \color{gray} (0.00)}} & \makecell{0.99 \\ \tiny{ \color{gray} (0.03)}} & \makecell{0.48 \\ \tiny{ \color{gray} (0.49)}} & \makecell{0.96 \\ \tiny{ \color{gray} (0.05)}} & \makecell{0.96 \\ \tiny{ \color{gray} (0.02)}} & \makecell{1.00 \\ \tiny{ \color{gray} (0.00)}} & \makecell{0.97 \\ \tiny{ \color{gray} (0.05)}} & \makecell{0.97 \\ \tiny{ \color{gray} (0.05)}} & \makecell{0.68 \\ \tiny{ \color{gray} (0.33)}} \\
 NearMiss                  & \makecell{1.00 \\ \tiny{ \color{gray} (0.00)}} & \makecell{1.00 \\ \tiny{ \color{gray} (0.00)}} & \makecell{1.00 \\ \tiny{ \color{gray} (0.00)}} & \makecell{0.90 \\ \tiny{ \color{gray} (0.30)}} & \makecell{0.98 \\ \tiny{ \color{gray} (0.05)}} & \makecell{1.00 \\ \tiny{ \color{gray} (0.00)}} & \makecell{1.00 \\ \tiny{ \color{gray} (0.00)}} & \makecell{1.00 \\ \tiny{ \color{gray} (0.00)}} & \makecell{0.49 \\ \tiny{ \color{gray} (0.30)}} & \makecell{1.00 \\ \tiny{ \color{gray} (0.00)}} \\
 TomekLinks                & \makecell{1.00 \\ \tiny{ \color{gray} (0.00)}} & \makecell{1.00 \\ \tiny{ \color{gray} (0.00)}} & \makecell{0.99 \\ \tiny{ \color{gray} (0.02)}} & \makecell{0.29 \\ \tiny{ \color{gray} (0.43)}} & \makecell{0.94 \\ \tiny{ \color{gray} (0.12)}} & \makecell{0.97 \\ \tiny{ \color{gray} (0.04)}} & \makecell{1.00 \\ \tiny{ \color{gray} (0.00)}} & \makecell{0.78 \\ \tiny{ \color{gray} (0.12)}} & \makecell{0.78 \\ \tiny{ \color{gray} (0.28)}} & \makecell{0.37 \\ \tiny{ \color{gray} (0.32)}} \\
 EditedNearestNeighbours   & \makecell{0.99 \\ \tiny{ \color{gray} (0.02)}} & \makecell{1.00 \\ \tiny{ \color{gray} (0.00)}} & \makecell{1.00 \\ \tiny{ \color{gray} (0.00)}} & \makecell{0.29 \\ \tiny{ \color{gray} (0.43)}} & \makecell{0.98 \\ \tiny{ \color{gray} (0.03)}} & \makecell{0.97 \\ \tiny{ \color{gray} (0.03)}} & \makecell{1.00 \\ \tiny{ \color{gray} (0.00)}} & \makecell{0.78 \\ \tiny{ \color{gray} (0.11)}} & \makecell{0.82 \\ \tiny{ \color{gray} (0.28)}} & \makecell{0.43 \\ \tiny{ \color{gray} (0.32)}} \\
 AllKNN                    & \makecell{1.00 \\ \tiny{ \color{gray} (0.00)}} & \makecell{1.00 \\ \tiny{ \color{gray} (0.00)}} & \makecell{1.00 \\ \tiny{ \color{gray} (0.00)}} & \makecell{0.29 \\ \tiny{ \color{gray} (0.43)}} & \makecell{0.98 \\ \tiny{ \color{gray} (0.03)}} & \makecell{0.96 \\ \tiny{ \color{gray} (0.04)}} & \makecell{1.00 \\ \tiny{ \color{gray} (0.00)}} & \makecell{0.78 \\ \tiny{ \color{gray} (0.11)}} & \makecell{0.82 \\ \tiny{ \color{gray} (0.28)}} & \makecell{0.43 \\ \tiny{ \color{gray} (0.32)}} \\
 OneSidedSelection         & \makecell{1.00 \\ \tiny{ \color{gray} (0.00)}} & \makecell{1.00 \\ \tiny{ \color{gray} (0.00)}} & \makecell{1.00 \\ \tiny{ \color{gray} (0.00)}} & \makecell{0.90 \\ \tiny{ \color{gray} (0.30)}} & \makecell{0.92 \\ \tiny{ \color{gray} (0.16)}} & \makecell{0.97 \\ \tiny{ \color{gray} (0.04)}} & \makecell{1.00 \\ \tiny{ \color{gray} (0.00)}} & \makecell{0.75 \\ \tiny{ \color{gray} (0.17)}} & \makecell{0.52 \\ \tiny{ \color{gray} (0.31)}} & \makecell{0.45 \\ \tiny{ \color{gray} (0.35)}} \\
 CondensedNearestNeighbour & \makecell{0.99 \\ \tiny{ \color{gray} (0.02)}} & \makecell{1.00 \\ \tiny{ \color{gray} (0.00)}} & \makecell{0.98 \\ \tiny{ \color{gray} (0.03)}} & \makecell{1.00 \\ \tiny{ \color{gray} (0.00)}} & \makecell{0.79 \\ \tiny{ \color{gray} (0.29)}} & \makecell{0.97 \\ \tiny{ \color{gray} (0.04)}} & \makecell{1.00 \\ \tiny{ \color{gray} (0.00)}} & \makecell{0.91 \\ \tiny{ \color{gray} (0.13)}} & \makecell{0.59 \\ \tiny{ \color{gray} (0.41)}} & \makecell{0.59 \\ \tiny{ \color{gray} (0.41)}} \\
 NeighbourhoodCleaningRule & \makecell{0.99 \\ \tiny{ \color{gray} (0.02)}} & \makecell{1.00 \\ \tiny{ \color{gray} (0.00)}} & \makecell{0.99 \\ \tiny{ \color{gray} (0.03)}} & \makecell{0.29 \\ \tiny{ \color{gray} (0.43)}} & \makecell{0.98 \\ \tiny{ \color{gray} (0.03)}} & \makecell{0.97 \\ \tiny{ \color{gray} (0.04)}} & \makecell{1.00 \\ \tiny{ \color{gray} (0.00)}} & \makecell{0.85 \\ \tiny{ \color{gray} (0.11)}} & \makecell{0.82 \\ \tiny{ \color{gray} (0.28)}} & \makecell{0.45 \\ \tiny{ \color{gray} (0.34)}} \\
\bottomrule
\end{tabularx}}
\end{adjustbox}
\end{table}

\begin{figure}[]
\centering
\begin{subfigure}{0.50\linewidth}
\includegraphics[width=\linewidth]{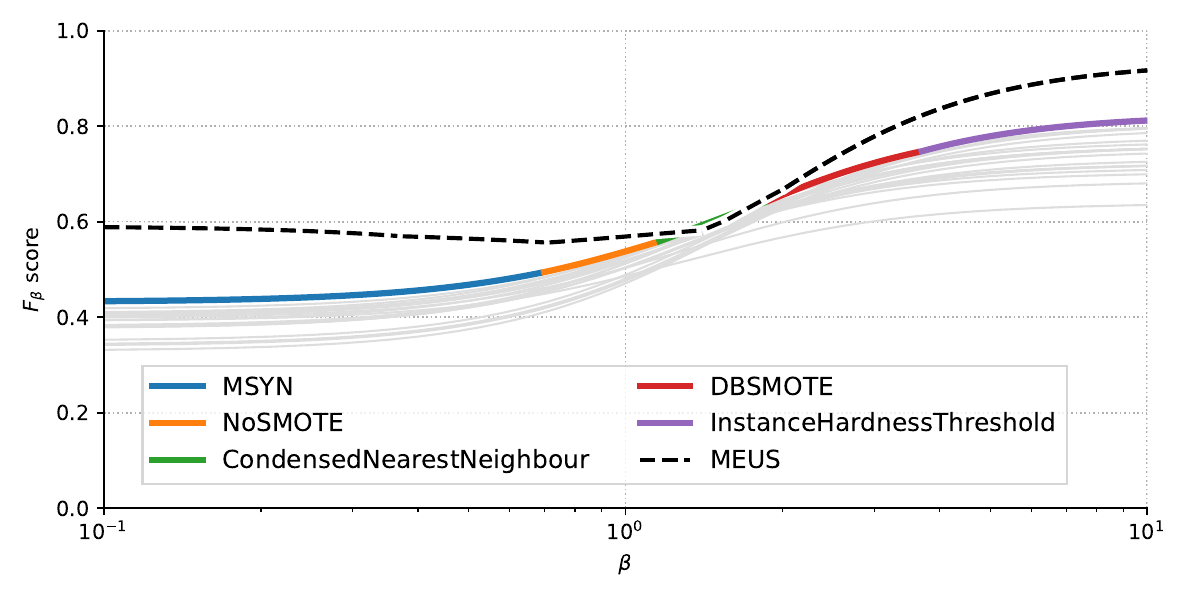} 
\caption{$DS_1$}
\end{subfigure}\hfill
\begin{subfigure}{0.50\linewidth}
\includegraphics[width=\linewidth]{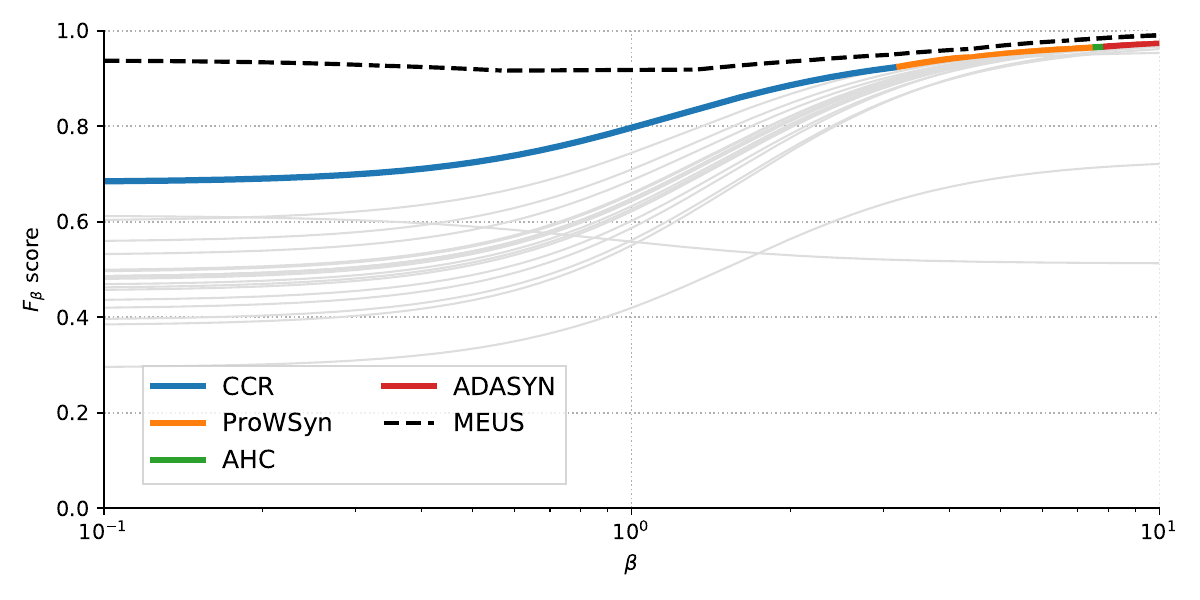} 
\caption{$DS_2$}
\end{subfigure}\hfill
\begin{subfigure}{0.49\linewidth}
\includegraphics[width=\linewidth]{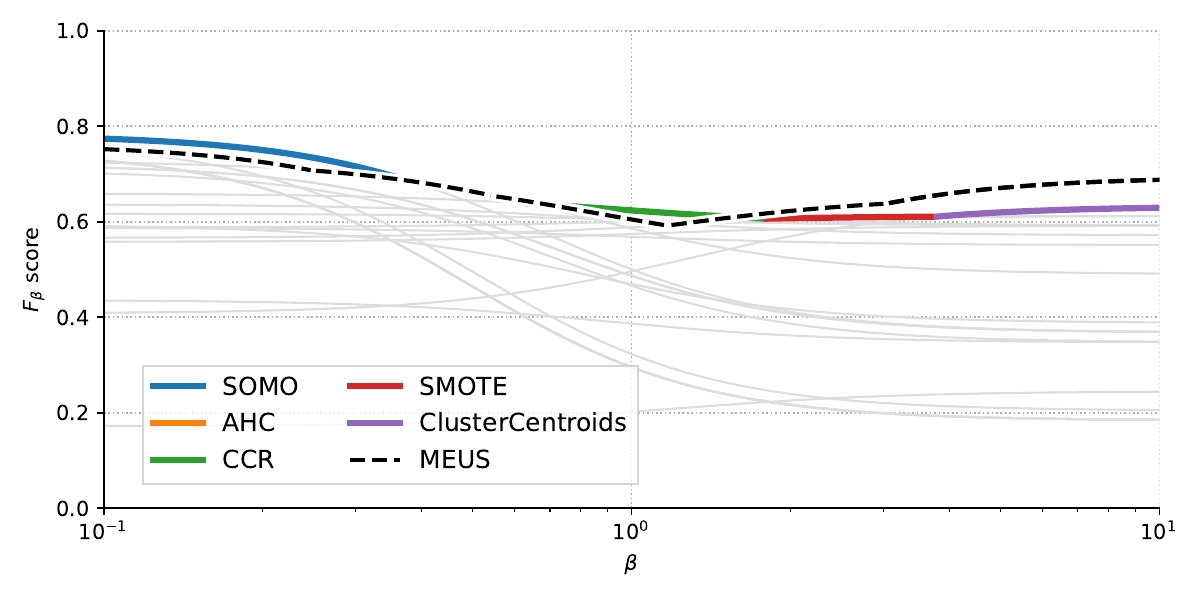} 
\caption{$DS_3$}
\end{subfigure}\hfill
\begin{subfigure}{0.49\linewidth}
\includegraphics[width=\linewidth]{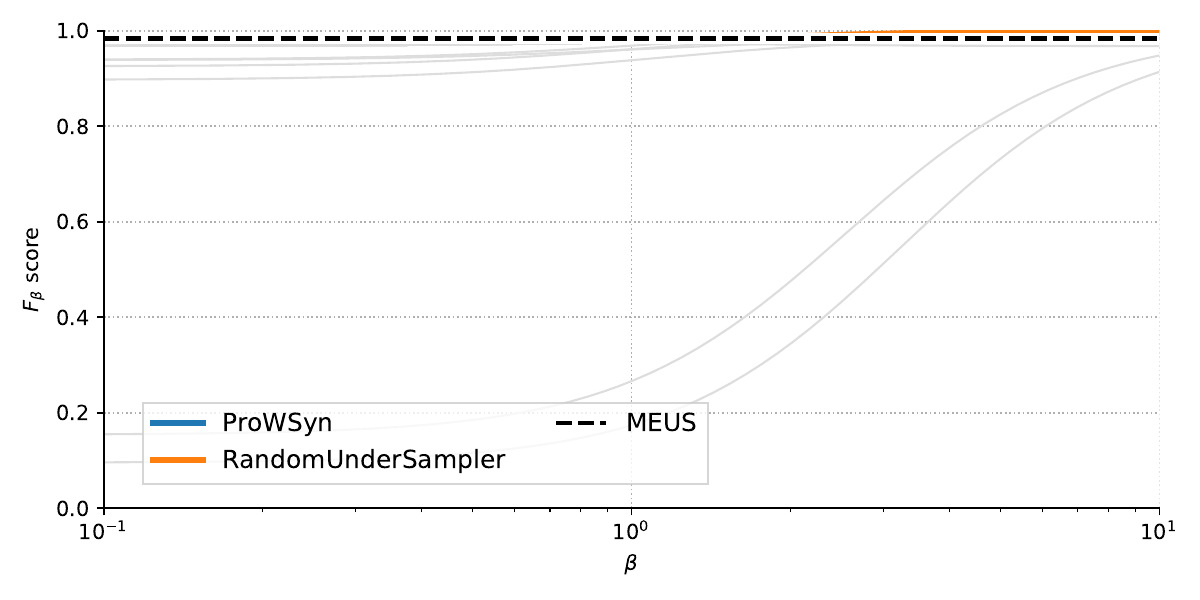} 
\caption{$DS_4$}
\end{subfigure}\hfill
\begin{subfigure}{0.49\linewidth}
\includegraphics[width=\linewidth]{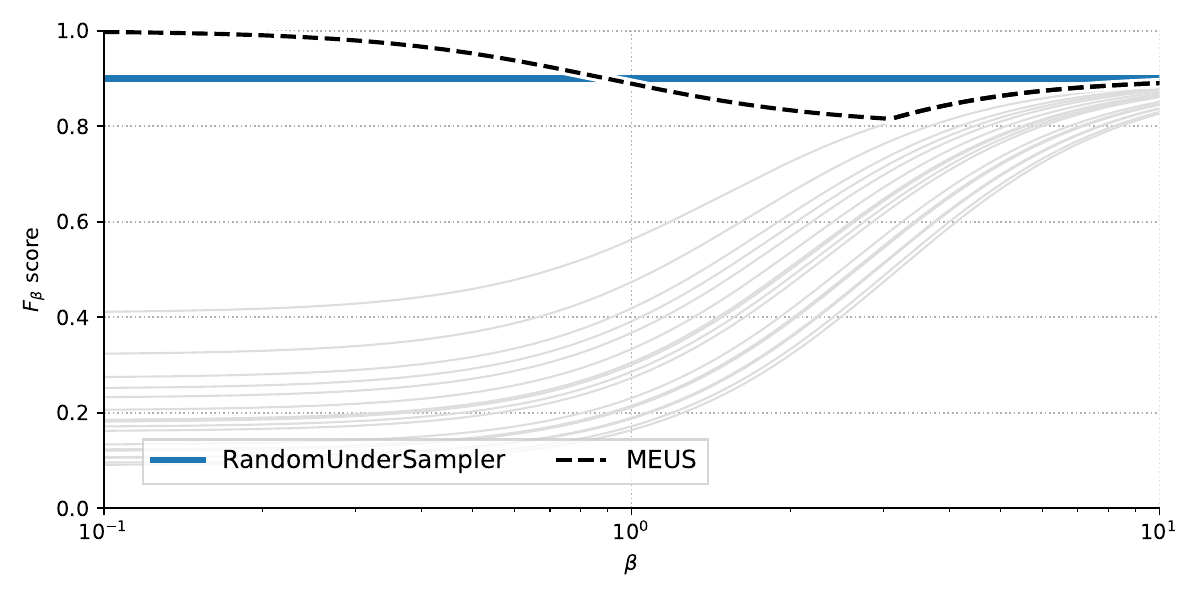} 
\caption{$DS_5$}
\end{subfigure}\hfill
\begin{subfigure}{0.49\linewidth}
\includegraphics[width=\linewidth]{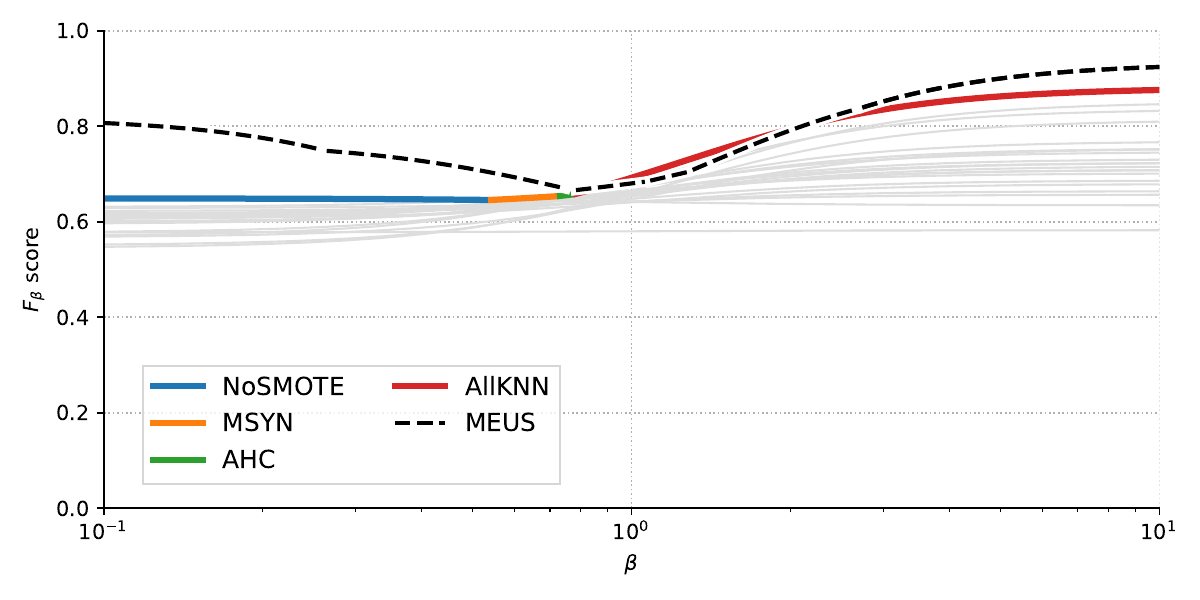} 
\caption{$DS_6$}
\end{subfigure}\hfill
\begin{subfigure}{0.49\linewidth}
\includegraphics[width=\linewidth]{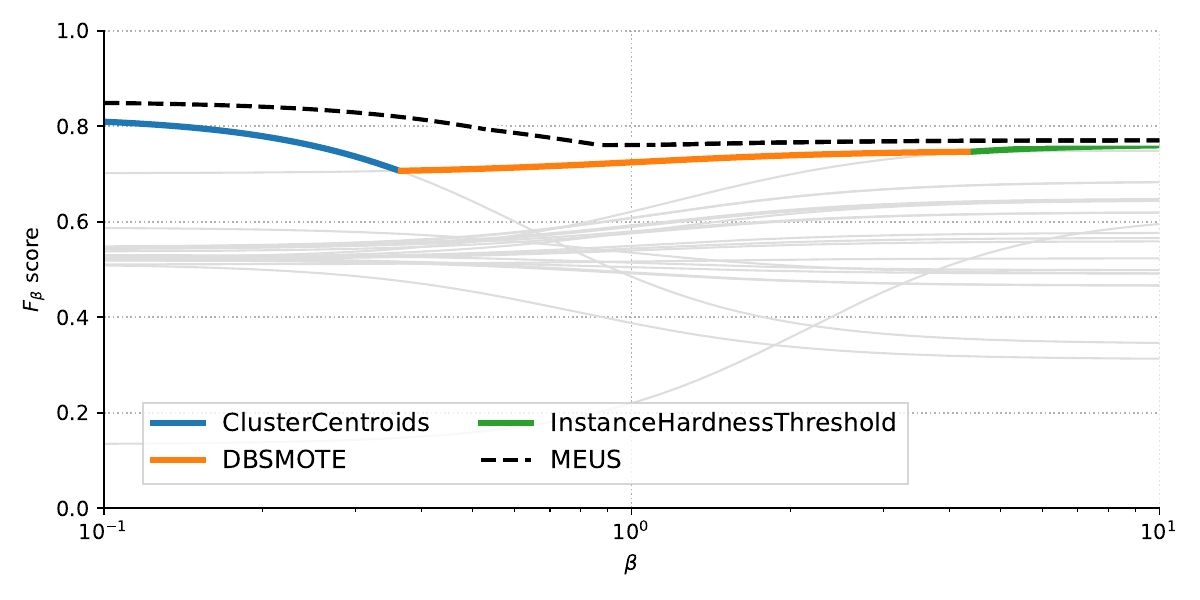} 
\caption{$DS_7$}
\end{subfigure}\hfill
\begin{subfigure}{0.49\linewidth}
\includegraphics[width=\linewidth]{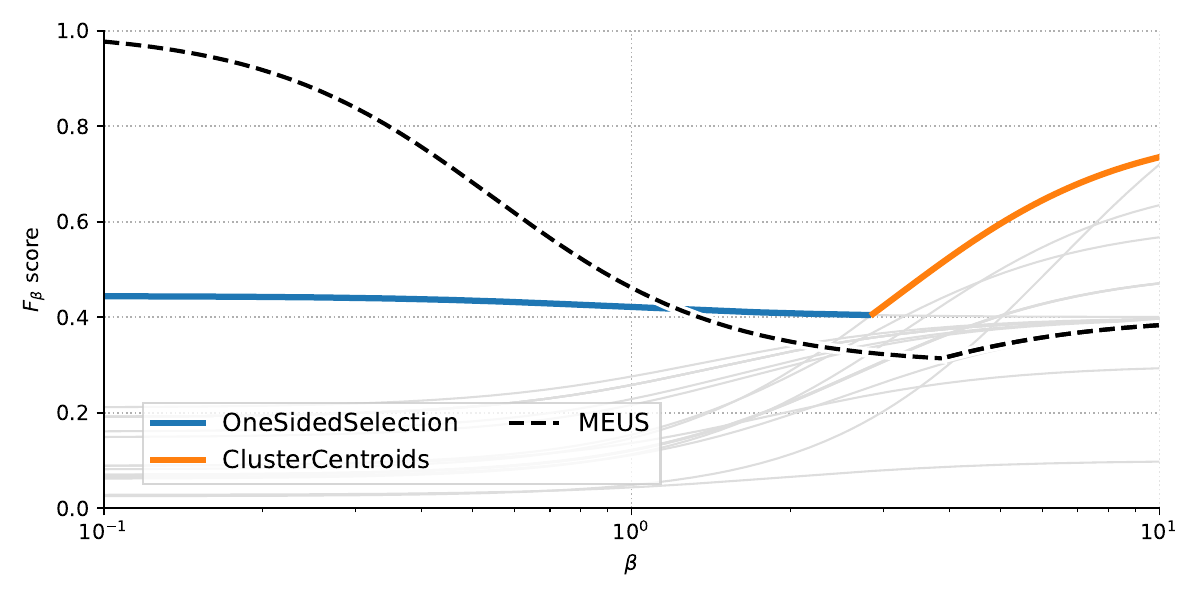} 
\caption{$DS_8$}
\end{subfigure}\hfill
\begin{subfigure}{0.49\linewidth}
\includegraphics[width=\linewidth]{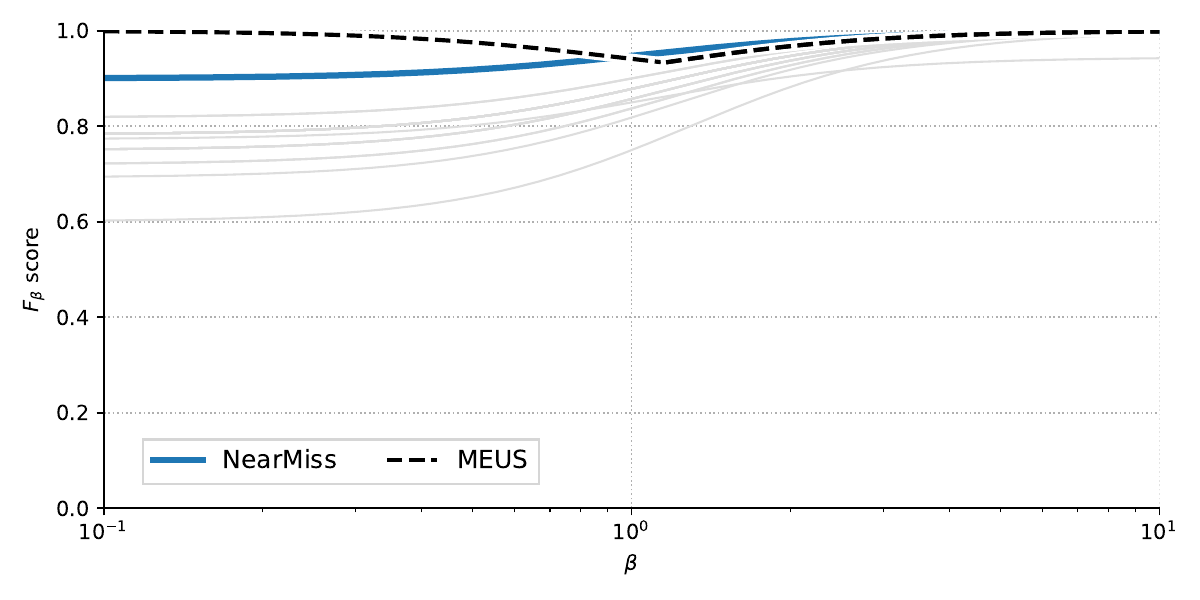} 
\caption{$DS_9$}
\end{subfigure}\hfill
\begin{subfigure}{0.49\linewidth}
\includegraphics[width=\linewidth]{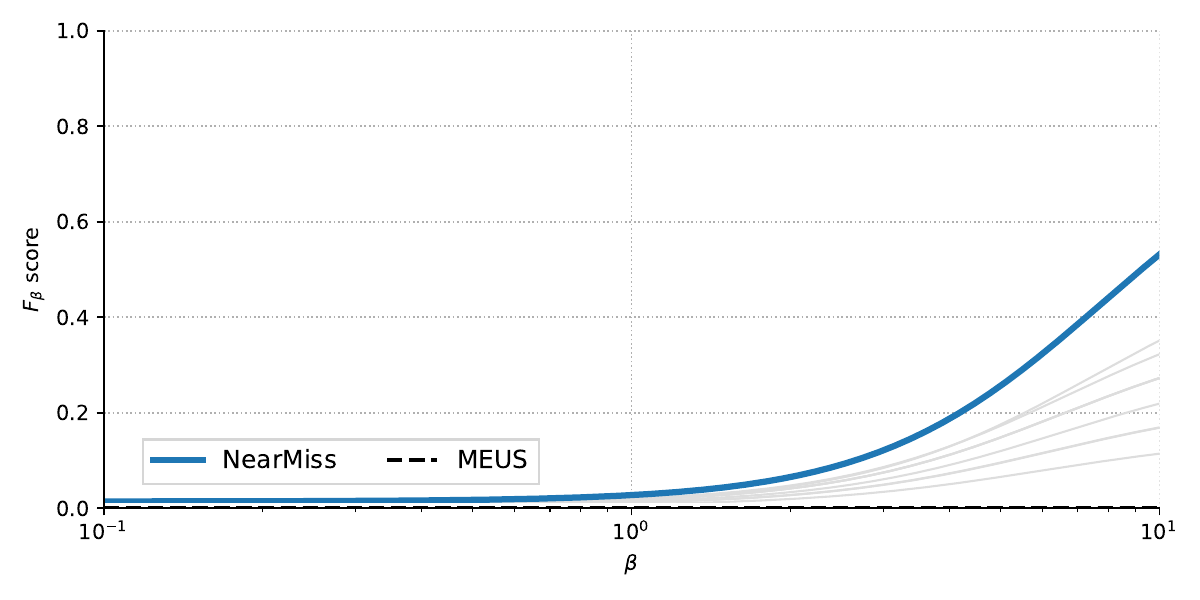} 
\caption{$DS_{10}$}
\end{subfigure}\hfill
\caption{$F_{\beta}$-plot results.}
\label{fig:fbeta}
\end{figure}

The values on the \textit{SDR} metric are comparable to those of the \textit{GD} metric and indicate an advantage over the NearMiss algorithm. It is challenging to consider the results stable due to the significant deviations in observed values. However, MEUS can often improve the results for problems $\mathcal{DS}_7$ and $\mathcal{DS}_2$.

With respect to the \textit{NDR} metric, it is evident that MEUS consistently attains remarkably high and stable values. However, when examining the data set, including weak-dominating solutions, it becomes evident that the proposed solution pool can provide diversity compared to the other resamplers. However, it is essential to note that exceptions such as MSYN in $\mathcal{DS}_4$ do exist. In the context of these observations, it is recommended that the average values be considered in favor of the reference method.

\subsection{Experiment B: Results}

The $F_{\beta}$ curves shown in Figure \ref{fig:fbeta} demonstrate the effectiveness of the selected methods in aligning with the user's preferences and the resultant quality of the chosen approach. The dashed line indicates the optimal model from the MEUS set. It can be seen that the \textsc{mol} algorithm outperforms the reference algorithms in $\mathcal{DS}_1$, $\mathcal{DS}_2$, and $\mathcal{DS}_3$. Notably, the set $\mathcal{DS}_2$ exhibits a very high level of consistency across all algorithms, with nearly all performing well and consistently settling on a high value, regardless of user preference. Furthermore, it can be observed that MEUS is not a suitable resampling method for the $\mathcal{DS}_{10}$ set, as was previously noted in the preceding experiment. Additionally, it is noteworthy that the previous experiment could not identify ClusterCentroids above average in $\mathcal{DS}_8$ and NearMiss in $\mathcal{DS}_{10}$. However, it should be noted that the results presented in the figures are not stabilized by cross-validation and may be influenced by the specifics of the chosen split.

\subsection{Discussion}

The principal question is whether the proposed metrics facilitate a comparison of the algorithms returning many solutions with the reference algorithms offering a single solution. As it has been demonstrated, they are feasible, although, as anticipated, they cannot fully describe the relationship between solutions.

The \textit{GD} and \textit{SDR} are unstable (exhibiting a substantial dependency on data splitting), and the conclusions based only on the average value may be misleading. \textit{HV} shows better stability, although it has the disadvantage of producing very low metric values, which are additionally challenging to correlate with the preference and diversity of the algorithm.

The most significant amount of information is derived from \textit{NDR}; however, these values may be interpreted as overly optimistic. The low values (the indicators in which the \textsc{mol} algorithm is less effective) should be mainly remarked on for analysis. Notably, the number of solutions generated by the \textit{SDR} and \textit{NDR} metrics is significantly influenced by the number of solutions provided by the algorithm. In particular, when \textit{NDR} is high and the pool of solutions is highly concentrated, the results may appear positively biased, as observed in the experiment.

In the case of the second experiment, it is evident that the methodology lacks stability from a single split approach. Obtaining this is quite challenging. While a statistical evaluation of the $F_{\beta}$-plot for the reference solutions is possible, the nature of the \textsc{mol} solutions would necessitate their interpretation as a whole. Nevertheless, in the case of a single split, the proposed method is highly effective in determining the \textsc{mol} algorithms' quality and the area in which they perform well. Furthermore, it demonstrates potential areas where a preference for shifting the \textsc{mol} algorithm would be advised.

\section{Conclusion}

To the best of our knowledge, this is the first work identifying a gap in the classifier learning evaluation regarding comparisons of methods returning multiple solutions with methods returning a single solution.

Despite emerging limitations for each proposed method, they facilitate the comparison and enable the characteristics of the \textsc{mol} model to be determined. It should be kept in mind that there are specific characteristics of methods (as shown in the case study). However, a conscious usage of them can provide more insights into the interpretation of the results and algorithm design.

It is advised to focus on the interpretation of results by visualizing relationships between solutions, such as $F_{\beta}$-plot visualization. Such an approach (similar to many explanation techniques) is not suitable for cross-validation, including statistical evaluations. However, in terms of the end-system user, this approach can help to find a best-fit solution tailored to his needs.

\section*{Acknowledgments}

This work was supported by the Polish National Science Centre under the grant No. 2019/35/B/ST6/04442.% and the Statutory fund of the Department of Systems and Computer Networks, Wrocław University of Science and Technology.

%Acknowledgments anonymized for the review.

\bibliographystyle{splncs04}
\bibliography{bibliography}

\end{document}